\documentclass{article}

\PassOptionsToPackage{numbers, compress}{natbib}
\usepackage[main, final]{neurips_2026}
\makeatletter
\renewcommand{\@noticestring}{%
}
\makeatother

\usepackage[utf8]{inputenc} 
\usepackage[T1]{fontenc}    
\usepackage{hyperref}       
\usepackage{url}            
\usepackage{booktabs}       
\usepackage{amsfonts}       
\usepackage{nicefrac}       
\usepackage{microtype}      
\usepackage{xcolor}         
\usepackage{graphicx}
\usepackage{subcaption}
\usepackage{array}
\usepackage{amsthm}
\usepackage{nicematrix}
\usepackage{tikz}
\usepackage{multirow}
\usepackage{threeparttable}
\usepackage{tablefootnote}
\usepackage{soul}
\usetikzlibrary{fit}
\usepackage{graphbox}
\usepackage{tikz}
\usepackage{adjustbox}
\usepackage{placeins}

\title{Concept Driven Domain Adaptation: Finding an Abstract Needle in a Haystack}

\author{%
  Haiming Zhao$^{1}$, Tai Wang$^{1}$\thanks{Corresponding Author}, Kun Zhang$^{1}$, Xicheng Peng$^{1}$, Zhiyang Li$^{2}$\\
  $^{1}$Laboratory for Artificial Intelligence and New Forms of Education \\
  Faculty of Artificial Intelligence in Education, Central China Normal University, Wuhan, China \\
  $^{2}$Wuhan United Imaging Healthcare Co., Ltd., Wuhan, China \\
  \texttt{\{wangtai, zhk, pengxicheng123\}@ccnu.edu.cn} \\
  \texttt{haimingzhao@mails.ccnu.edu.cn} \\
  \texttt{zhiyang.li@united-imaging.com} \\
}

\begin{document}

\maketitle

\begin{abstract}
Science teachers frequently search for documentary excerpts not by describing what appears on screen, but by querying the abstract concepts they intend to teach. This use case exposes a limitation of existing language-based video moment retrieval methods, which typically assume that queries describe observable events, whereas instructional search requires retrieving concrete visual phenomena that instantiate an underlying scientific principle. We study this setting as \emph{concept-to-example video retrieval}, an abstract-needle-in-a-haystack problem where compact curriculum concepts must be grounded in temporally sparse documentary evidence. To bridge this abstraction gap, we propose Concept-Driven Domain Adaptation (CDDA), a three-stage framework for adapting two-tower vision--language models to concept-level retrieval. CDDA treats concepts as intermediate semantic anchors: it first structures the textual embedding space with textbook and teacher-handbook example--concept pairs, then transfers this concept-aware geometry to documentary visuals under a frozen visual encoder, and finally jointly adapts both encoders with sparse visual concept supervision. From a geometric perspective, this staged alignment reduces text--concept and vision--concept angular gaps, thereby encouraging concept-level adaptation while preserving the pretrained model's concrete image--description alignment. On a curated middle-school physics retrieval benchmark, CDDA achieves stronger pedagogically oriented concept retrieval than several competitive multimodal baselines, including Qwen3-VL-Embedding-2B, while maintaining concrete image--text matching after adaptation.
\end{abstract}

\section{Introduction}

Science education often requires students to connect abstract principles with concrete phenomena. Concepts such as inertia, thermal expansion, and universal gravitation are difficult to grasp from definitions alone, but become meaningful when students can observe relevant instances, reason about them, and relate them to formal knowledge. Short video clips are well suited to this process, as they make otherwise invisible or inaccessible phenomena observable and support classroom explanation, discussion, and demonstration. Science documentaries offer a particularly useful source of such material, combining rich visual evidence, expert narration, and scientific exposition. However, their instructional use is constrained by a practical retrieval bottleneck: teachers must manually search through long-form videos to locate brief segments that illustrate a target curriculum concept. Prior work shows that videos are widely used in instruction \cite{Kaltura2018}, that teachers prefer concise clips aligned with learning objectives \cite{Boclips2019,Pattier2021}, and that shorter videos can improve engagement and learning outcomes \cite{Zhu2022}. These findings motivate a need not directly addressed by existing retrieval systems: retrieving short, scientifically credible documentary segments from abstract curriculum concepts.

This setting departs from the assumptions of standard video--text retrieval and temporal grounding. Existing methods typically take concrete natural-language queries that describe the desired moment\cite{xiao2024bridging,huang2024vtimellm,deng2025seq2time,ren2024timechat}, such as ``a ball falls from a tower'' or ``the Moon orbits the Earth,'' and retrieve segments that are visually or linguistically aligned with the description. Educational search, however, often starts from the concept to be taught rather than a scene description. A segment relevant to universal gravitation may show planetary motion, falling bodies, tides, or historical demonstrations, even if the term itself is never mentioned. Relevance therefore depends less on lexical overlap or caption-like correspondence than on whether the segment provides a phenomenon that can be explained by the queried principle.

We formulate this problem as concept-to-example video retrieval: given a curriculum concept and a collection of long-form science documentaries, retrieve short segments that visually instantiate the concept. The task introduces an abstraction mismatch largely absent from conventional benchmarks: queries are compact, abstract, and explanatory, whereas the evidence is concrete, multimodal, temporally sparse, and often indirectly related to the query words\cite{hannan2025revisionllm,shen2024tempme}. This abstract needle-in-a-haystack problem is amplified in documentaries\cite{li2024test}, which are organized around narratives, discoveries, historical figures, and visually engaging events rather than textbook taxonomies. For example, a documentary about Newton may interleave mechanics, optics, astronomy, biography, and historical context; the most useful teaching segment may be embedded in this narrative and described through visible events rather than curriculum terminology. Models trained mainly to match captions to video moments are therefore not optimized for the educational relation of interest\cite{lin2024mm,xie2025smartclip,hsu2025makes}: whether a visual segment can serve as an explanatory example of an abstract scientific concept.

To address this challenge, we propose Concept-Driven Domain Adaptation (CDDA), a staged framework for adapting two-tower vision--language models to concept-to-example retrieval. CDDA is based on the observation that the required alignment is difficult to learn from sparse video annotations alone: direct fine-tuning on documentary segment--concept pairs can associate curriculum labels with observed videos, but provides little structure for why diverse visual segments and textual examples instantiate the same principle. CDDA separates adaptation into three stages. First, we adapt the language encoder with example--concept pairs from physics textbooks and teachers' handbooks, shaping the text space so that concepts are close to the explanations and examples they govern. Second, we align this curriculum-aware language space with a frozen visual encoder using sparse documentary segment--concept annotations, transferring concept structure to the visual domain while reducing overfitting. Third, we jointly fine-tune both encoders on cross-modal concept supervision to fit the documentary domain after the concept space has been structured. This design encourages concepts, textual examples, and visual segments to align through shared scientific meaning rather than lexical overlap or caption-level similarity.

Our contributions are threefold. First, we identify and formalize concept-to-example video retrieval, where abstract curriculum concepts are used to retrieve concrete visual phenomena from long-form science documentaries. Second, we introduce CDDA, a concept-driven adaptation framework that combines textbook-derived textual supervision with sparse documentary annotations to bridge abstract scientific concepts and multimodal visual evidence. Third, we show that concept-driven alignment improves abstract concept retrieval over the original CLIP model and a strong off-the-shelf multimodal embedding baseline, while preserving performance on concrete image--text matching.

\section{Related works}
\paragraph{Language-based video moment retrieval.}
Language-based video moment retrieval encompasses temporal sentence grounding, natural language video localization, video moment retrieval, and corpus-level moment retrieval. These tasks aim to localize or retrieve video moments that match a natural-language query, either within an untrimmed video or from a larger video corpus. Existing methods are commonly organized around how candidate moments are represented and selected: proposal-based methods rank predefined temporal segments\cite{Radford2021}, proposal-free methods directly regress moment boundaries\cite{zhang2020span,zhang2020learning}, and recent query-based or DETR-style models learn moment proposals through cross-modal interaction\cite{lei2021detecting,moon2023query,sun2024tr,lee2024bam,xiao2024bridging}. Recent work \cite{dong2024temporal,guo2025vtg,hannan2025revisionllm,chen2024verified}further extends this paradigm to query absence, video LLM reasoning, and long-form grounding . Despite these advances, the semantic contract remains largely unchanged: the query describes what should be observed. Our setting changes this contract. The query is a curriculum-level principle, while the target segment is an observable phenomenon that instantiates or is governed by that principle. Thus, the central challenge is not merely retrieving or localizing moments, but bridging an abstraction gap between explanatory concepts and concrete multimodal evidence.

\paragraph{Adapting vision--language models.}
Large-scale vision--language models, especially two-tower models such as CLIP, provide a strong foundation for efficient cross-modal retrieval through contrastive image--text pretraining \cite{Radford2021}. Prior work adapts such models to downstream tasks using prompt tuning, adapters, parameter-efficient fine-tuning, or task-specific alignment modules \cite{Hu2022,Yang2024,Luo2023}. Recent studies further show that adaptation may introduce task- or data-specific misalignment, motivating mechanisms that preserve general visual-language knowledge while improving target-domain performance \cite{zhang2024rethinking}. However, most adaptation methods assume that downstream supervision remains on the same semantic plane as pretraining, such as matching images with captions, attributes, class names, or domain-specific descriptions. In concept-to-example retrieval, the textual label is not a paraphrase of the visual content but a higher-level explanation of it. Directly fine-tuning on sparse video--concept pairs can therefore fit associations without organizing the underlying principle--phenomenon relation. CDDA addresses this by first shaping the language space with textual example--concept supervision, then transferring this curriculum-aware structure to visual segments through staged cross-modal adaptation.

\paragraph{Abstract concept grounding.}
Abstract concepts are difficult for multimodal models because they often lack stable visual appearances and cannot be grounded through object-level recognition alone. Prior work has studied abstract visual understanding through metaphorical image retrieval, abstract image similarity, and structured concept grounding \cite{Cerini2024,Tater2024,hsu2025makes}. Schema-based approaches further show that decomposing abstract concepts into intermediate components can help VLMs reason beyond literal visual correspondence \cite{hsu2025makes}. Our work shares the view that abstract grounding requires additional conceptual structure, but studies a different form of abstraction. Scientific concepts such as gravity, inertia, and thermal expansion are abstract not because they are subjective or metaphorical, but because they denote latent principles whose instances appear as observable phenomena. This distinguishes our task from knowledge-augmented captioning or entity enrichment, which mainly relies on definitional, taxonomic, or descriptive relations \cite{Shen2022,Sharma2018}. Concept-to-example video retrieval instead requires aligning explanatory principles with the concrete visual evidence they govern, motivating the use of educational texts as intermediate supervision for cross-modal adaptation.

\section{Method}
\label{sec:method}

\subsection{Problem Formulation}

We consider concept-to-example video retrieval from long-form science documentaries. Let
$\mathcal{D}=\{D_m\}_{m=1}^{M}$ denote a collection of documentaries, which are segmented into a pool of candidate clips
$\mathcal{V}=\{v_i\}_{i=1}^{N_v}$. Let
$\mathcal{C}=\{c_y\}_{y=1}^{N_c}$ be a set of curriculum concepts, where each concept is represented by its name and, when available, a short definition. Given a concept query $c_y$, the goal is to rank all clips $v\in\mathcal{V}$ by their usefulness as concrete visual examples of the queried concept:
\begin{equation}
    S(v,c_y): \mathcal{V}\times \mathcal{C}\rightarrow \mathbb{R},
\end{equation}
where a larger score indicates stronger concept-level relevance.

Unlike conventional text-video retrieval, the query in our setting is not necessarily a literal description of the target visual content. A relevant clip may instantiate the scientific principle behind the query without explicitly mentioning the concept name. We therefore use two complementary sources of supervision. The first is a textual concept corpus
$\mathcal{T}=\{(u_i,y_i)\}_{i=1}^{N_t}$, where $u_i$ is a textual example, phenomenon description, or explanatory passage, and $y_i$ denotes the concept it instantiates. The second is a sparse visual annotation set
$\mathcal{A}=\{(v_i,y_i)\}_{i=1}^{N_a}$, where $v_i$ is a documentary clip labeled as an instructional example of concept $c_{y_i}$.

\begin{figure*}[t]
    \centering
    \scriptsize
    \captionsetup[subfigure]{font=scriptsize,skip=1pt}
    \captionsetup[figure]{font=small,skip=3pt}

    \begin{subfigure}[c]{0.46\textwidth}
        \centering
        \includegraphics[width=\linewidth]{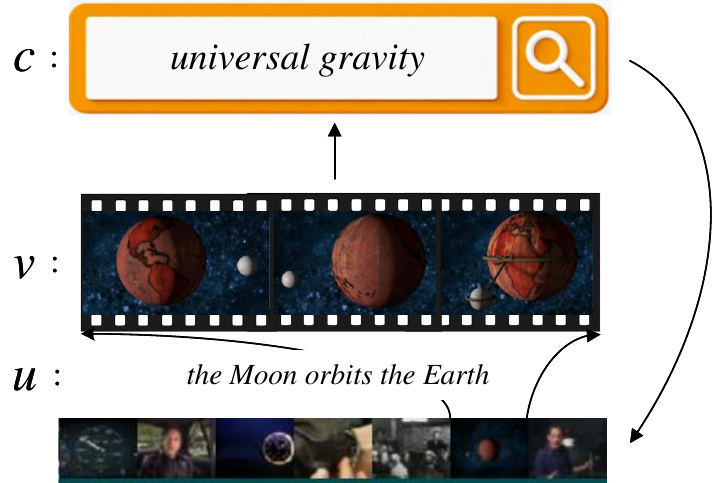}
        \caption{Concept-to-example video retrieval.}
        \label{fig:frame_series}
    \end{subfigure}
    \hspace{10pt}
    \begin{subfigure}[c]{0.27\textwidth}
        \centering
        \includegraphics[width=\linewidth]{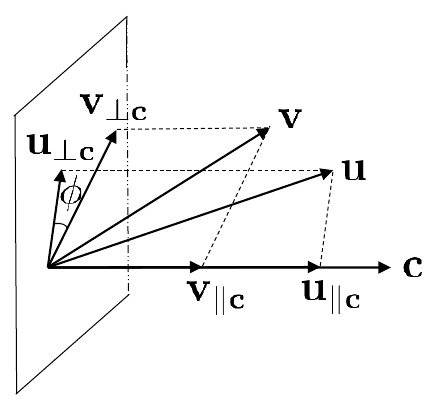}
        \caption{Geometric intuition.}
        \label{fig:vectors}
    \end{subfigure}

    \vspace{-5pt}
    \caption{
    Illustration of concept-to-example video retrieval and its geometric motivation.
    Left: an abstract curriculum concept is used as a query to retrieve concrete documentary clips that instantiate the underlying scientific principle.
    Right: the concept embedding serves as an intermediate semantic anchor between concrete textual examples and visual instances.
    }
    \label{fig:concept_geometry}
\end{figure*}

\subsection{Model Overview}

We build CDDA on a two-tower vision--language model with a visual encoder $f_{\theta}$ and a text encoder $g_{\phi}$. For a video clip $v$, we sample $K$ frames or short sub-clips, encode them with the visual encoder, and aggregate them with a temporal pooling operator:
\begin{equation}
    \mathbf{z}_v =
    \frac{
    h_v\!\left(\mathrm{Pool}(f_{\theta}(v))\right)
    }{
    \left\|h_v\!\left(\mathrm{Pool}(f_{\theta}(v))\right)\right\|_2
    },
\end{equation}
where $h_v$ is a visual projection head. For any text input $x$, including either a concept $c_y$ or a textual example $u_i$, we compute
\begin{equation}
    \mathbf{z}_x =
    \frac{
    h_t(g_{\phi}(x))
    }{
    \left\|h_t(g_{\phi}(x))\right\|_2
    },
\end{equation}
where $h_t$ is a textual projection head. Retrieval is performed by cosine similarity:
\begin{equation}
    S(v,c_y)=\mathbf{z}_v^{\top}\mathbf{z}_{c_y}.
\end{equation}

\begin{figure}[t]
    \centering
    \includegraphics[width=0.85\linewidth]{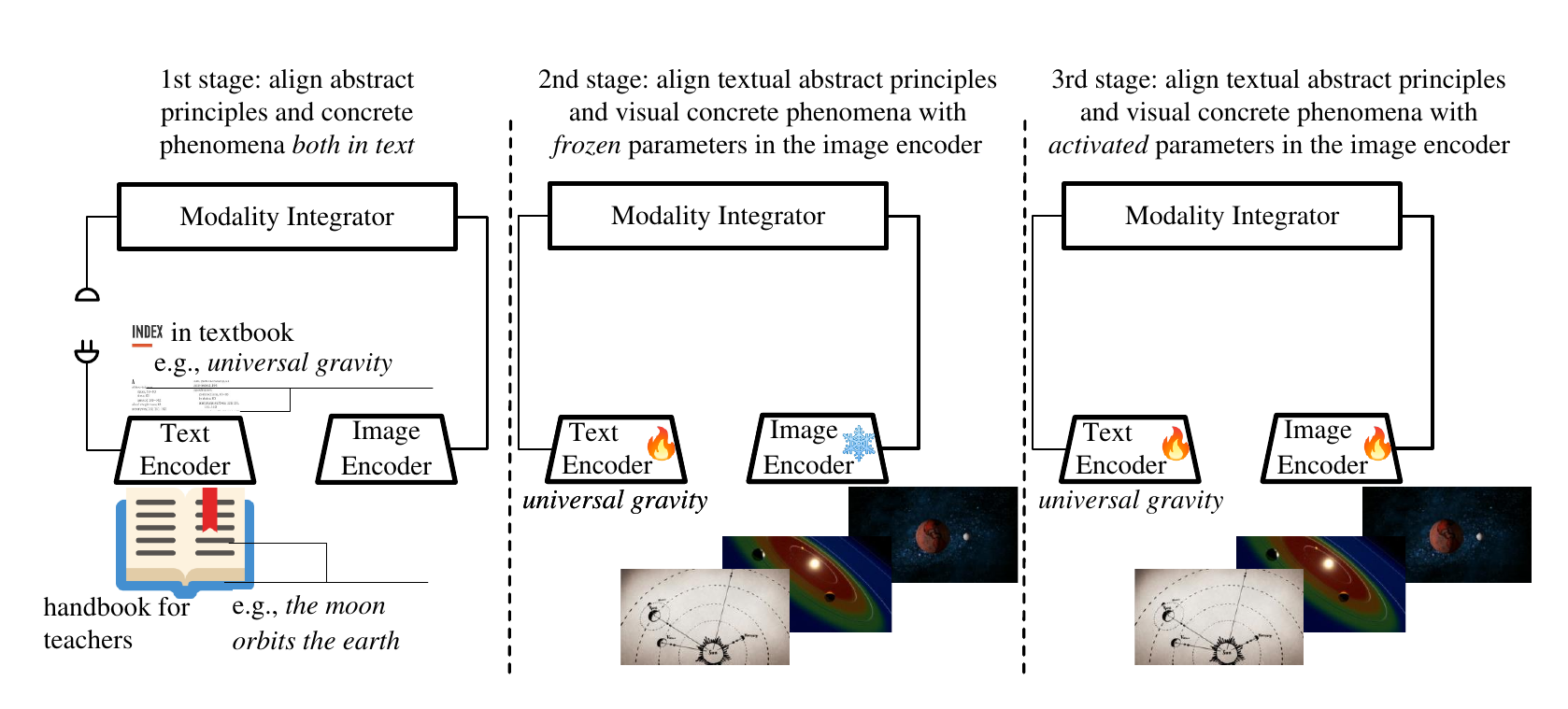}
    \caption{
    Overview of CDDA. Stage 1 structures the text space using example--concept pairs from educational texts. Stage 2 aligns documentary clips with the curriculum-aware concept space while keeping the visual encoder frozen. Stage 3 jointly adapts both encoders to the documentary domain.
    }
    \label{fig:stages}
\end{figure}

As shown in Fig.~\ref{fig:stages},CDDA adapts this model in three stages. Stage 1 structures the language space using textual example--concept pairs. Stage 2 transfers the resulting concept structure to video clips while keeping the visual backbone frozen. Stage 3 jointly adapts the full model to the documentary domain with textual replay.

\subsection{Design Rationale: Concepts as Semantic Anchors}
\label{the:consistrans}

Before introducing the training objectives, we motivate CDDA from a geometric perspective: curriculum concepts can serve as intermediate semantic anchors between concrete textual examples and visual instances in the normalized embedding space.Let
\begin{equation}
    \theta(a,b)=\arccos(\mathbf{z}_a^\top\mathbf{z}_b)
\end{equation}
denote the angular distance between two embeddings on the unit hypersphere. For a video clip $v$, a concrete textual example $u$, and a concept $c$, the spherical triangle inequality gives
\begin{equation}
    \theta(v,u) \leq \theta(v,c) + \theta(u,c).
\end{equation}
Thus, reducing the text--concept distance $\theta(u,c)$ in Stage 1 and the video--concept distance $\theta(v,c)$ in Stage 2 decreases an upper bound on the video--example distance. As illustrated in Fig.~\ref{fig:concept_geometry}, the concept embedding acts as an intermediate semantic anchor between textual examples and visual instances.
This argument does not claim that concept supervision alone guarantees video-text alignment. Rather, it explains why directly fine-tuning on sparse video--concept pairs may be less stable than first constructing a curriculum-aware concept space and then transferring it to video. More specifically, we have 
\begin{equation}
\cos \theta_{\mathbf{v}\mathbf{u}} = \cos \theta_{\mathbf{v}\mathbf{c}}\cos \theta_{\mathbf{u}\mathbf{c}}+\sin \theta_{\mathbf{v}\mathbf{c}}\sin \theta_{\mathbf{u}\mathbf{c}}\cos \phi
\end{equation}
The formal proof and additional discussion are provided in the appendix.

\subsection{Multi-Positive Concept Alignment}

Both textual examples and video clips may have many-to-one relations with curriculum concepts: multiple descriptions or clips can instantiate the same scientific principle. We therefore use a multi-positive concept alignment objective instead of a one-to-one contrastive objective.

For a mini-batch
$\mathcal{B}=\{(a_i,y_i)\}_{i=1}^{B}$,
where $a_i$ can be either a textual example or a video clip, let
$\mathcal{Y}=\{y_i:(a_i,y_i)\in\mathcal{B}\}$ be the set of concepts appearing in the batch. We define the prototype of concept $y$ as the normalized embedding of its textual concept representation:
\begin{equation}
    \mathbf{p}_y = \mathbf{z}_{c_y}.
\end{equation}
The instance-to-concept loss is
\begin{equation}
\mathcal{L}_{a\rightarrow c}
=
-\frac{1}{B}
\sum_{i=1}^{B}
\log
\frac{
\exp(\mathbf{z}_{a_i}^{\top}\mathbf{p}_{y_i}/\tau)
}{
\sum_{y\in\mathcal{Y}}
\exp(\mathbf{z}_{a_i}^{\top}\mathbf{p}_{y}/\tau)
},
\end{equation}
where $\tau$ is a temperature parameter. The reverse direction treats all instances of the same concept in the batch as positives:
\begin{equation}
\mathcal{L}_{c\rightarrow a}
=
-\frac{1}{|\mathcal{Y}|}
\sum_{y\in\mathcal{Y}}
\log
\frac{
\sum_{i:y_i=y}
\exp(\mathbf{p}_{y}^{\top}\mathbf{z}_{a_i}/\tau)
}{
\sum_{i=1}^{B}
\exp(\mathbf{p}_{y}^{\top}\mathbf{z}_{a_i}/\tau)
}.
\end{equation}
The resulting alignment objective is
\begin{equation}
    \mathcal{L}_{\mathrm{mp}}(\mathcal{B};\tau)
    =
    \frac{1}{2}
    \left(
    \mathcal{L}_{a\rightarrow c}
    +
    \mathcal{L}_{c\rightarrow a}
    \right).
\end{equation}
When each concept appears only once in a mini-batch, this objective reduces to the standard symmetric contrastive form. When multiple instances share the same concept, it avoids incorrectly pushing them apart as negatives.

\subsection{Stage 1: Concept-Text Structuring}

The first stage adapts the language space using the textual corpus $\mathcal{T}$. For a mini-batch
$\mathcal{B}_t=\{(u_i,y_i)\}_{i=1}^{B_t}$, we minimize
\begin{equation}
    \mathcal{L}_{\mathrm{text}}
    =
    \mathcal{L}_{\mathrm{mp}}(\mathcal{B}_t;\tau_t).
\end{equation}
This stage updates the text encoder and textual projection head. Its purpose is to make abstract concept embeddings close to the textual examples, explanations, and phenomenon descriptions that instantiate them. The visual encoder is not used in this stage.

\subsection{Stage 2: Video-Concept Alignment}

The second stage transfers the structured concept space to documentary clips. For a visual mini-batch
$\mathcal{B}_v=\{(v_i,y_i)\}_{i=1}^{B_v}$, we optimize
\begin{equation}
    \mathcal{L}_{\mathrm{video}}
    =
    \mathcal{L}_{\mathrm{mp}}(\mathcal{B}_v;\tau_v).
\end{equation}
During this stage, the visual backbone $f_{\theta}$ is frozen, while the projection heads and text-side parameters are updated. Freezing the visual backbone provides a stable visual anchor and reduces overfitting under sparse video supervision. Concept prototypes $\mathbf{p}_y$ are computed from the Stage-1 text space, so video clips are aligned to curriculum-aware concept representations rather than to unstructured concept labels.

\subsection{Stage 3: Joint Concept-Driven Adaptation}

The final stage unfreezes both encoders and adapts the full model to the documentary domain. We optimize
\begin{equation}
    \mathcal{L}_{\mathrm{joint}}
    =
    \mathcal{L}_{\mathrm{video}}
    +
    \eta \mathcal{L}_{\mathrm{text}},
\end{equation}
where $\eta$ controls the strength of textual replay. The video loss adapts the model to visual instances of scientific concepts, while textual replay regularizes the concept space learned in Stage 1. This allows CDDA to benefit from end-to-end cross-modal adaptation without discarding the concept--example structure induced from educational texts. The reason why freezing was necessary was that \cite{Hao2026} reported visible
evidence showing that there are “severe gradient conflicts between vision and text”. Such
conflicts can undermine training unless the text gradient is synchronized with the changes of vision gradient  prior to joint encoder training.

\subsection{Inference}

At inference time, CDDA requires only a concept query and the candidate video clips. We pre-compute embeddings for all clips in $\mathcal{V}$. Given a concept query $c_y$, we compute $\mathbf{z}_{c_y}$ and rank clips by
\begin{equation}
    S(v,c_y)=\mathbf{z}_v^{\top}\mathbf{z}_{c_y}.
\end{equation}
The top-ranked clips are returned as visual examples of the queried scientific concept. Adjacent high-ranking clips can be merged into longer excerpts for classroom use.


\section{Experiments}
\label{sec:experiments}

We evaluate CDDA on concept-to-example retrieval from long-form science documentaries. All methods are tested on the same 146 candidate segments, five curriculum concept queries, and expert-verified relevance labels. The experiments examine whether concept-driven adaptation improves abstract concept retrieval, whether it preserves concrete image--description matching, and which stages are responsible for the gains.

\subsection{Experimental Setup}
\label{subsec:exp_setup}

\paragraph{Benchmark.}
Because existing video-text benchmarks do not annotate visual evidence by curriculum concepts, we curate a middle-school physics benchmark centered on the force unit. Table~\ref{tab:source_material} lists the documentary sources used to build the image-concept supervision set; the held-out retrieval corpus is a different documentary and is never used for adaptation.

\begin{table}[t]
\centering
\small
\caption{Documentary sources used to construct the image-concept supervision set. ``Episodes'' denotes the number of episodes used in our dataset, not the total number of episodes in the original series.}
\label{tab:source_material}
\begin{tabular}{p{0.38\linewidth}c c p{0.25\linewidth}}
\toprule
Collection & Episodes & Duration & Production team \\
\midrule
When We Left Earth: The NASA Missions & 4 & 45 min. & Discovery Channel \\
Earth: The Power of the Planet & 1 & 60 min. & BBC Two \\
The Fabric of the Cosmos & 2 & 55 min. & PBS NOVA \\
100 Greatest Discoveries & 1 & 44 min. & Discovery Channel \\
Cosmos: A Spacetime Odyssey & 1 & 45 min. & National Geographic \\
What on Earth Is Wrong with Gravity? & 1 & 44 min. & BBC Horizon \\
\bottomrule
\end{tabular}
\end{table}

\begin{figure}[t]
\centering
\scriptsize
\setlength{\tabcolsep}{0pt}
\renewcommand{\arraystretch}{0.85}
\captionsetup[subfigure]{font=scriptsize,skip=1pt,justification=centering}
\begin{tabular}{@{}c@{\hspace{6pt}}c@{\hspace{10pt}}c@{\hspace{10pt}}c@{}}
\rotatebox{90}{\makebox[1.65cm][c]{\bfseries\emph{gravity}}} &
\subcaptionbox{An astronaut experiences weightlessness in a plane.}[2.55cm]{\includegraphics[height=1.55cm,keepaspectratio]{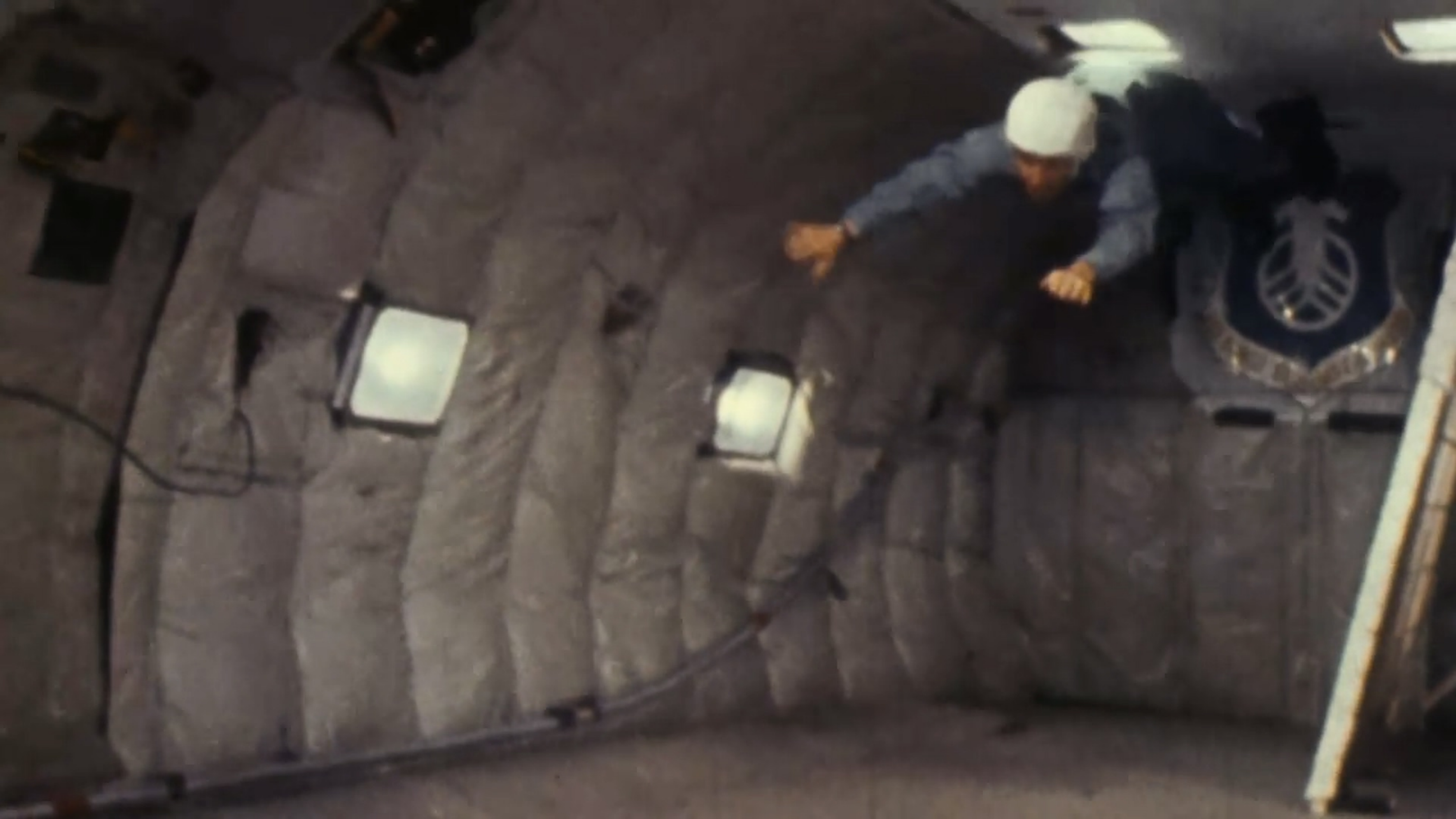}} &
\subcaptionbox{An astronaut drives a lunar rover on the Moon.}[2.55cm]{\includegraphics[height=1.55cm,keepaspectratio]{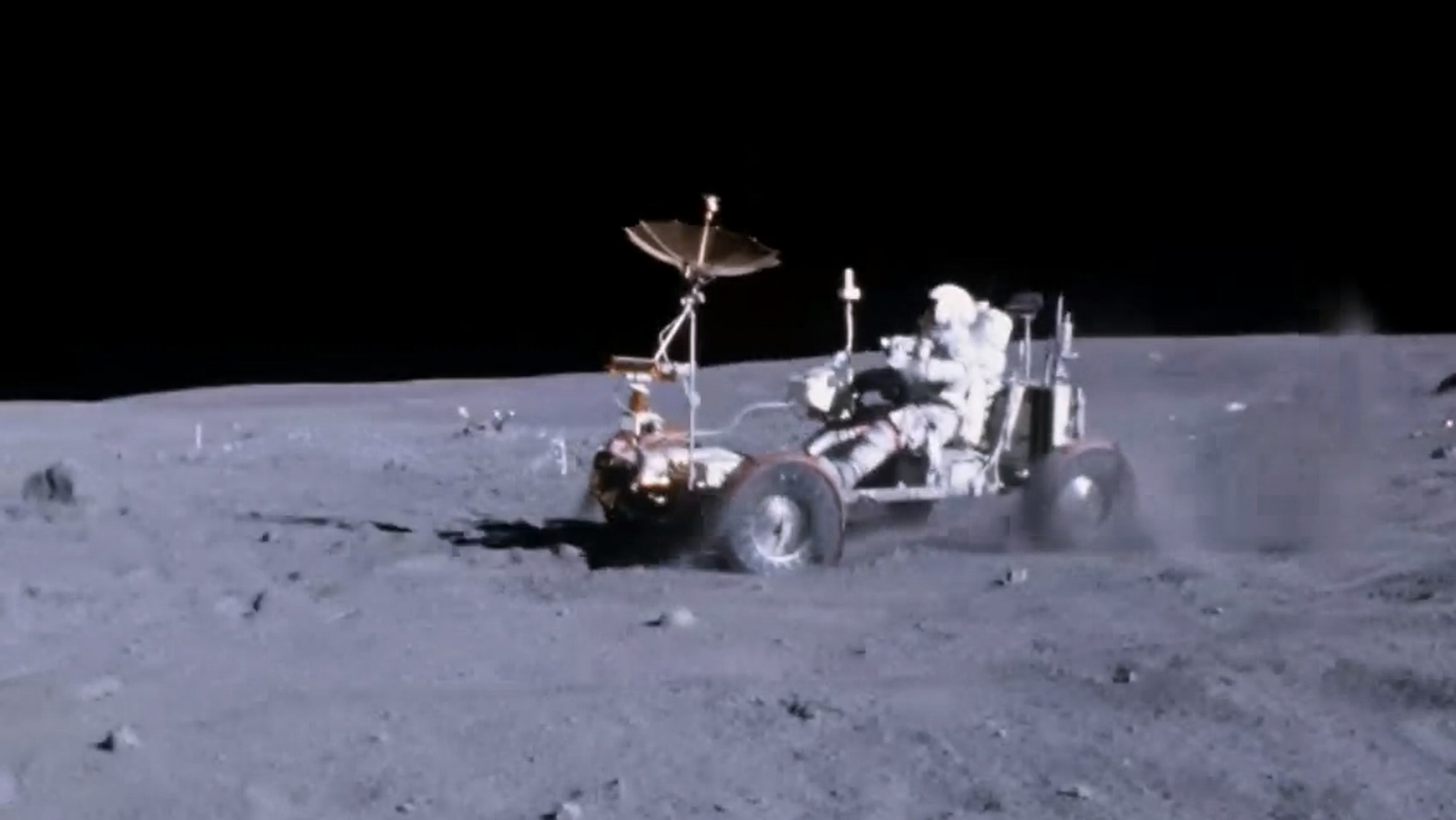}} &
\subcaptionbox{An astronaut experiences weightlessness in the Station.}[2.55cm]{\includegraphics[height=1.55cm,keepaspectratio]{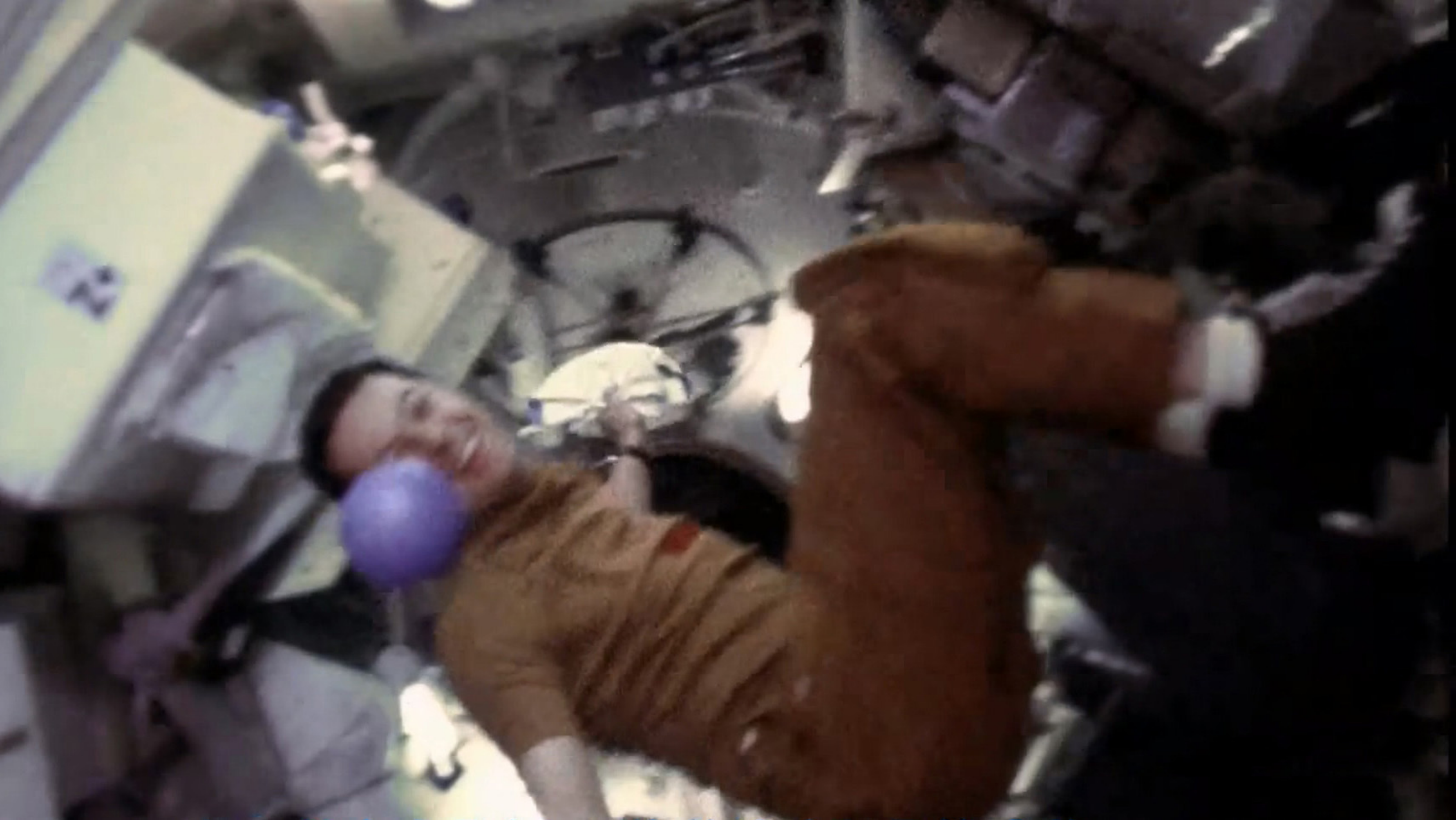}} \\
[3pt]
\rotatebox{90}{\makebox[1.65cm][c]{\shortstack[c]{\bfseries\emph{universal}\\[-1pt]\bfseries\emph{gravity}}}} &
\subcaptionbox{Planets orbit the Sun in circular orbits.}[2.55cm]{\includegraphics[height=1.55cm,keepaspectratio]{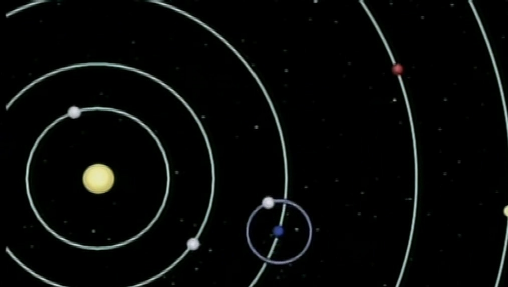}} &
\subcaptionbox{Planets orbit the Sun in elliptical orbits.}[2.55cm]{\includegraphics[height=1.55cm,keepaspectratio]{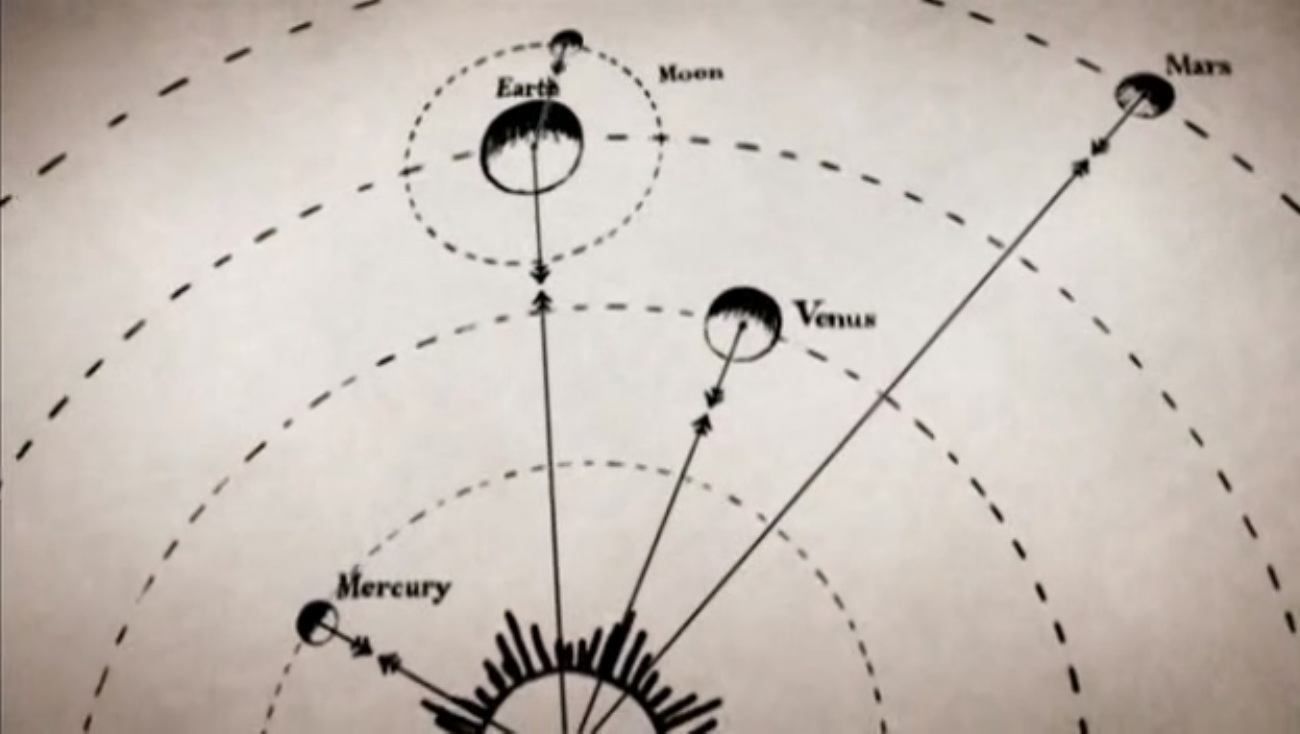}} &
\subcaptionbox{Planets orbit the Sun in elliptical orbits.}[2.55cm]{\includegraphics[height=1.55cm,keepaspectratio]{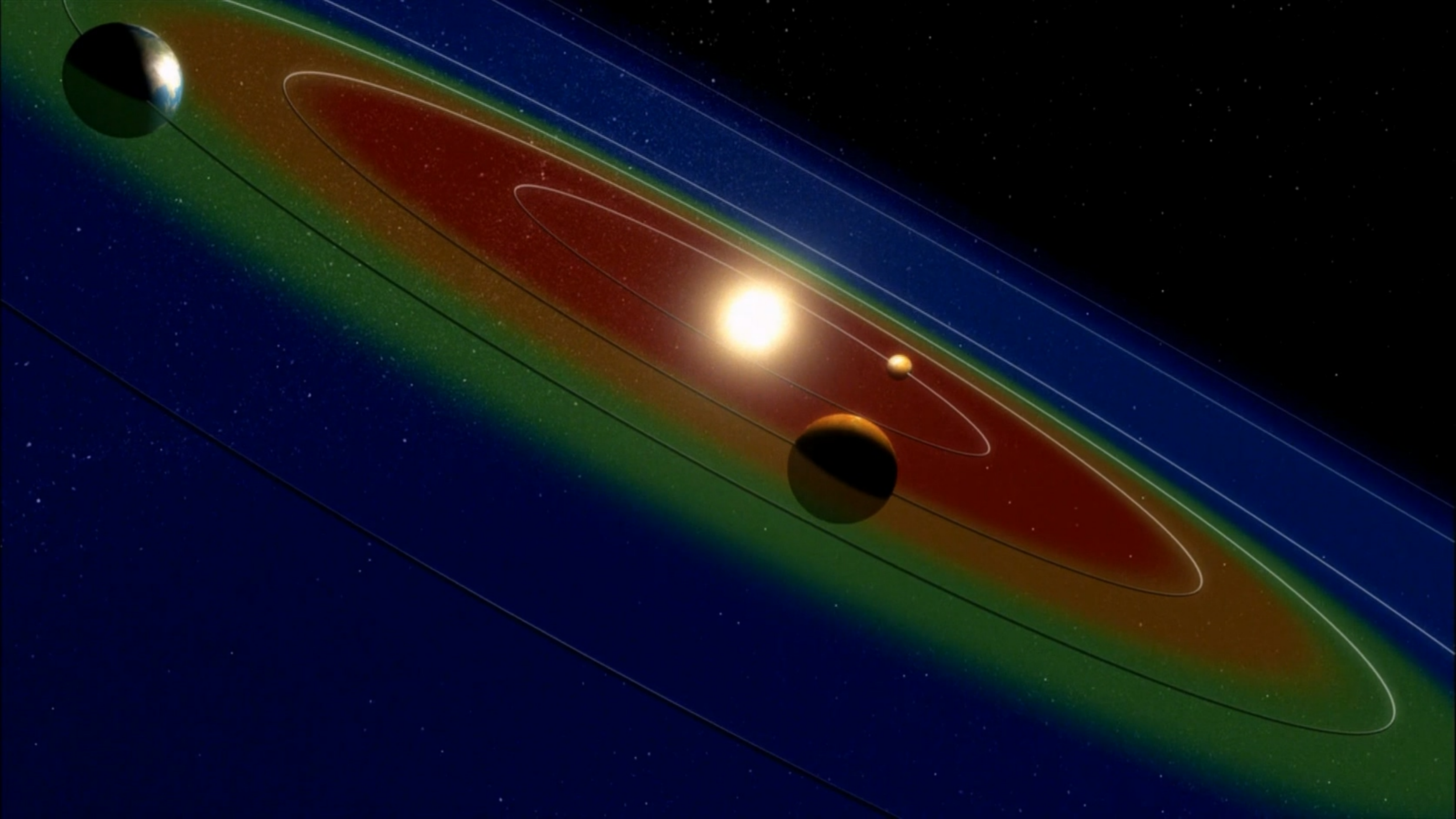}}
\end{tabular}
\vspace{-3pt}
\caption{Examples of visual concept supervision. Each image is paired with an abstract curriculum concept $c$; concrete descriptions $u$ are shown only for readability.}
\label{fig:visual_concept_supervision}
\end{figure}

The supervision set contains 62 image-level examples from Table~\ref{tab:source_material} and Motion Mountain (Vol.~1), split into 44/12/6 train/validation/test examples. The five concepts are Force, Gravity, Universal Gravity, Newton, and Newton's First Law of Motion. Labels indicate the curriculum concept that an image can instantiate in instruction, rather than a literal caption. Stage~1 further uses 150 textual example--concept pairs: 118 sentences from a teachers' handbook and 32 descriptions of the collected visual examples. Fig.~\ref{fig:visual_concept_supervision} shows the resulting many-to-one supervision pattern.

\paragraph{Retrieval corpus and models.}
Evaluation uses Newton's Dark Secrets, a 52.5-minute documentary with 832 subtitles. Following subtitle-embedding segmentation~\cite{Ghinassi2024}, we construct 146 non-overlapping candidate segments before model comparison, and only the segment embedding model varies across methods. CDDA is initialized from CN-CLIP~\cite{Yang2023}; the text-side adaptation component uses Chinese RoBERTa~\cite{Cui2021}. We compare with CN-CLIP~\cite{Yang2023}, Gemini Embedding~2~\cite{Lee2025}, Qwen3-VL-Embedding-2B~\cite{Li2026}, ImageBind~\cite{Girdhar2023}, InternVideo-Next~\cite{Wang2026}, and text-only retrieval over subtitles using BM25~\cite{Kamphuis2020}, CN-CLIP\textsubscript{text}~\cite{Cui2021}, CDDA(II\&III)$^{d}$\textsubscript{text}, and KBERT~\cite{Liu2020}. InternVideo-Next is evaluated as a visual-retrieval diagnostic by replacing each concept query with a representative concept image; the query images and implementation details are provided in Appendix~\ref{app:exp_details}.

\paragraph{Metrics.}
Two senior pedagogical experts cross-check relevance labels for the five queries. We report (i) top-10 pedagogical hits, which count whether returned segments are valid teaching examples, and (ii) MAP@K and NDCG@K for $K\in\{5,10\}$ with binary relevance. The non-trivial match (NtM) in Fig.~\ref{fig:ntm_variant} receives additional pedagogical credit only for Universal Gravity; for MAP/NDCG it is treated as an ordinary relevant segment. The marker ``almost'' indicates that a method retrieves the segment immediately before a relevant segment; these cases are shown for interpretability and are not counted in the technical metrics.

\subsection{Main Results}
\label{subsec:main_results}

\begin{table}[htbp]
\centering
\caption{Total hits of multiple large models on each abstract concept in top 10 results}
\begin{NiceTabular}{ l c l c c l c }
\toprule
                & \multirow{2}{*}{Force} & \multirow{2}{*}{Gravity} & \multirow{2}{*}{Newton}
                                         & Newton's  & Universal & pedagogical \\
                &       &         &      & First Law & Gravity   & scores      \\
\# of segments  & (6)   & (20)    & (9)  &    (1)    &   (25)    &              \\
\midrule

\textbf{CDDA}                         & \textbf{5} & \textbf{6}   & \textbf{5} & \textbf{almost} & \textbf{5+1NtM} & \textbf{24} \\
\quad CDDA(II\&III)                   & 3 & 6      & 4  & almost & 4+1NtM  & 20 \\
\quad CDDA(II\&III)\textsuperscript{d+c}
                                      & 1 & 5      & 3  & almost & 4       & 13 \\
\quad CDDA(II\&III)\textsuperscript{d}
                                      & 0 & 4+1    & 1  & almost & 3+1NtM  & 12 \\
\quad CDDA(II\&III)\textsuperscript{d}\_text
                                      & 1 & 5      & 1  & 0      & 5       & 12 \\

\midrule
Gemini Embed. 2 \cite{Lee2025}       & 2 & 8+1    & 0  & 1      & 9+1NtM  & 24 \\
Qwen3-VL-2B \cite{Li2026}            & 1 & 7      & 6  & 1      & 7       & 22 \\
ImageBind \cite{Girdhar2023}         & 2 & 6      & 5  & 0      & 8       & 21 \\
InternVideo-Next \cite{Wang2026}     & 1 & 5+1    & 5  & 0      & 8       & 20 \\
CN-CLIP \cite{Yang2023}              & 3 & 3+1    & 1  & almost & 4+1NtM  & 15 \\
CN-CLIP\_text \cite{Cui2021}         & 0 & 4      & 1  & 0      & 7       & 12 \\
BM25 \cite{Kamphuis2020}             & 0 & 5      & 2  & 1      & 2       & 10 \\
KBERT \cite{Liu2020}                 & 0 & 3      & 2  & 0      & 4       & 9  \\
\bottomrule
\end{NiceTabular}
\label{tb:dig_res}
\end{table}

Table~\ref{tb:dig_res} shows that CDDA ties Gemini Embedding~2 in pedagogical score and outperforms the open or trainable baselines under the same candidate pool. Its clearest advantage is on Force: CDDA retrieves 5 of 6 relevant segments, whereas general-purpose multimodal baselines retrieve at most 2. This concept is visually heterogeneous, covering gears, levers, balances, thrust, and pulling, and therefore benefits from concept-level alignment rather than surface visual templates. CDDA also retrieves the NtM segment for Universal Gravity without over-crediting it under Gravity, indicating better concept specificity.

\begin{table}[t]
\centering
\small
\caption{Standard retrieval metrics averaged over the five concept queries. Relevance is binary; the NtM segment is treated as an ordinary relevant segment for MAP and NDCG.}
\label{tab:ranking_metrics}
\begin{tabular}{lrrrr}
\toprule
Method & MAP@10 & NDCG@10 & MAP@5 & NDCG@5 \\
\midrule
Gemini Embed. 2 & 0.4924 & 0.5710 & 0.5300 & 0.5968 \\
\textbf{CDDA} & \textbf{0.3848} & \textbf{0.5402} & \textbf{0.3807} & \textbf{0.5094} \\
Qwen3-VL-2B & 0.3723 & 0.5374 & 0.3793 & 0.4814 \\
InternVideo-Next & 0.3245 & 0.4448 & 0.4167 & 0.4941 \\
CDDA(II\&III) & 0.2974 & 0.4309 & 0.3753 & 0.4601 \\
ImageBind & 0.2789 & 0.4161 & 0.2340 & 0.3470 \\
\bottomrule
\end{tabular}
\end{table}

Table~\ref{tab:ranking_metrics} confirms the same trend under standard ranking metrics. Gemini Embedding~2 obtains the strongest MAP values, but CDDA remains competitive despite using a smaller CN-CLIP backbone. Compared with Qwen3-VL-2B, CDDA improves MAP@10 from 0.3723 to 0.3848 and NDCG@5 from 0.4814 to 0.5094; compared with ImageBind, it improves MAP@10 by 0.1059 and NDCG@10 by 0.1241. Thus, concept-driven adaptation improves early ranking quality, not only the presence of relevant segments in the top-10 list.

\subsection{Diagnostics and Ablations}
\label{subsec:diagnostics_ablations}

\paragraph{Preservation of concrete alignment.}
Concept adaptation should not destroy the pretrained model's ability to match concrete images with literal descriptions. Table~\ref{tab:preservation} evaluates this diagnostic on the six held-out images. CDDA preserves the same image--description correctness pattern as the original CN-CLIP while improving several image--concept decisions. This supports the geometric motivation that curriculum-level alignment can be added without erasing concrete visual-textual correspondence.

\begin{table*}[t]
\centering
\scriptsize
\setlength{\tabcolsep}{3pt}
\renewcommand{\arraystretch}{0.86}
\caption{Preservation of image--description matching after concept adaptation. Each entry reports image--description correctness, followed by image--concept correctness in parentheses.}
\label{tab:preservation}
\begin{threeparttable}
\begin{adjustbox}{max width=0.98\textwidth,center}
\begin{NiceTabular}{c m{4.0cm}<{\centering} c c c c}
\toprule
Test image & Description (Concept) & CN-CLIP & CDDA & CDDA (I\&II) & Qwen3-VL-2B \\
\midrule
\includegraphics[height=0.82cm,align=c,keepaspectratio]{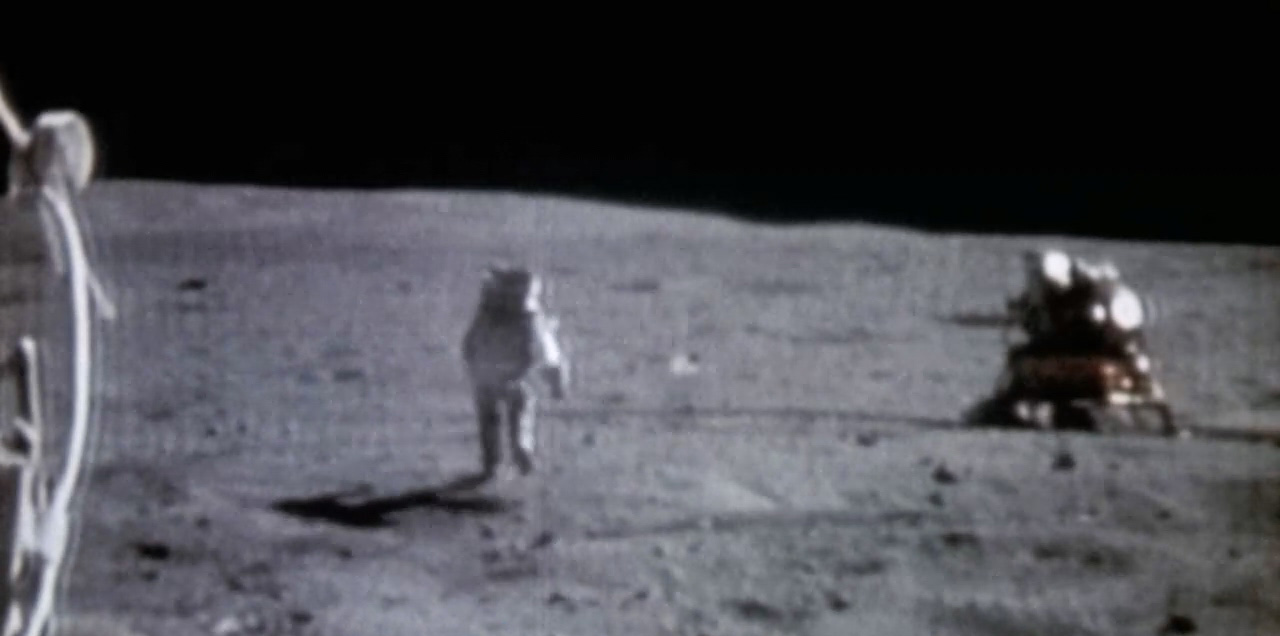} & An astronaut bounce-walks on the Moon. (Gravity) & $\checkmark$ ($\times$) & $\checkmark$ ($\checkmark$) & $\checkmark$ ($\checkmark$) & $\checkmark$ ($\times$) \\
\includegraphics[height=0.82cm,align=c,keepaspectratio]{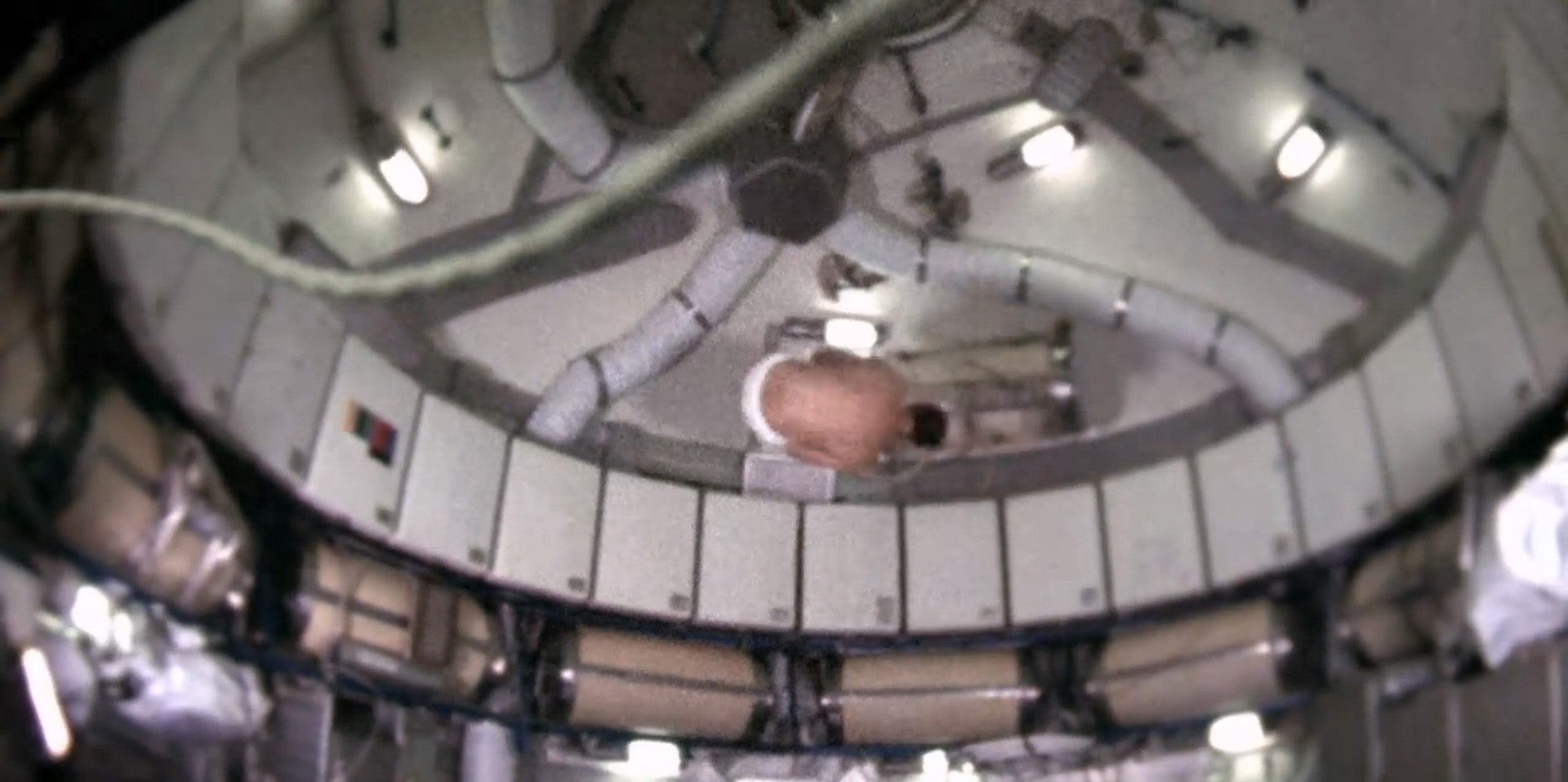} & An astronaut experiences weightlessness in the Station. (Gravity) & $\checkmark$ ($\checkmark$) & $\checkmark$ ($\checkmark$) & $\checkmark$ ($\checkmark$) & $\checkmark$ ($\times$) \\
\includegraphics[height=0.82cm,align=c,keepaspectratio]{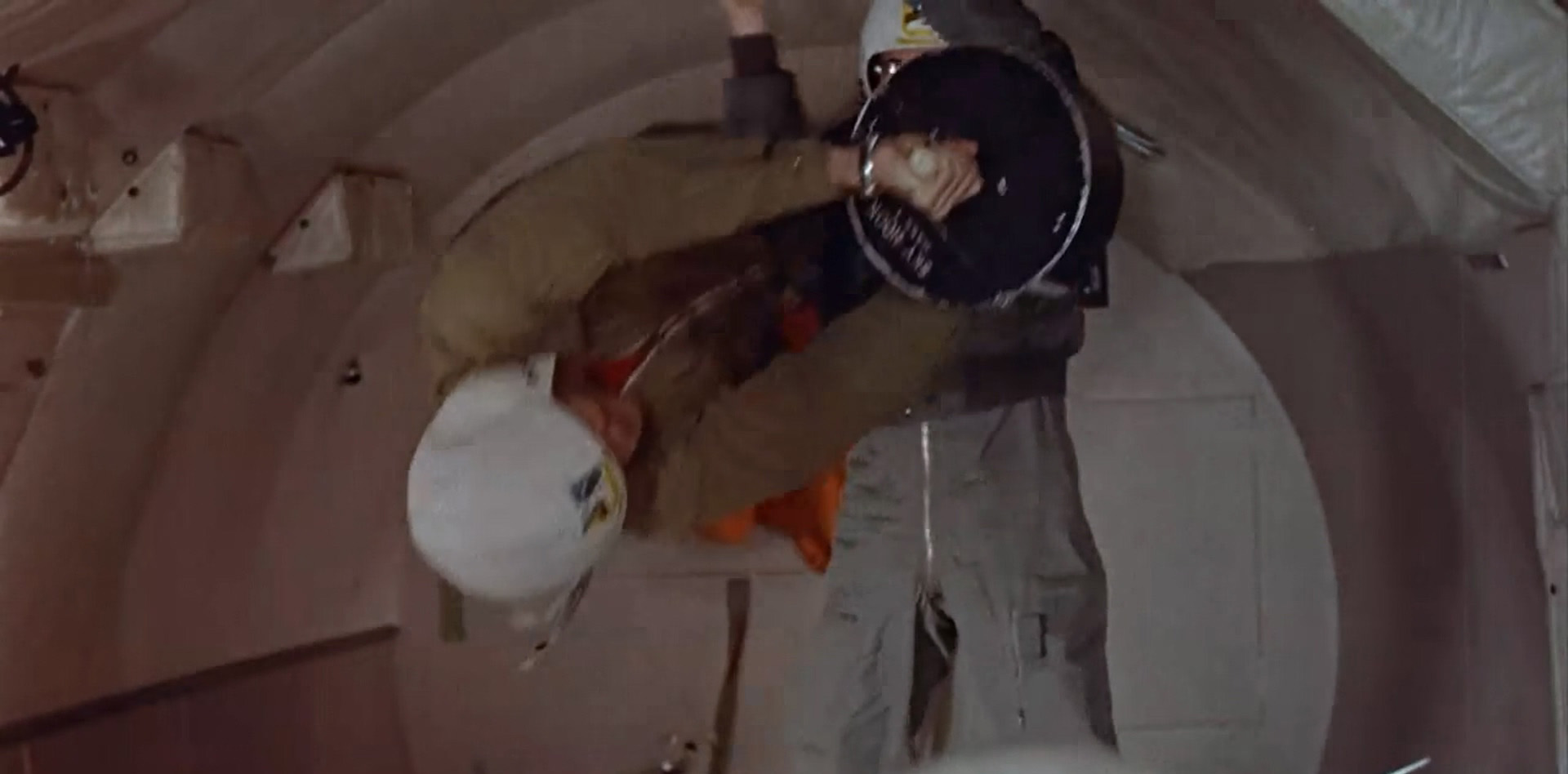} & An astronaut experiences weightlessness in a plane. (Gravity) & $\times$ ($\checkmark$) & $\times$ ($\checkmark$) & $\times$ ($\checkmark$) & $\times$ ($\checkmark$) \\
\includegraphics[height=0.82cm,align=c,keepaspectratio]{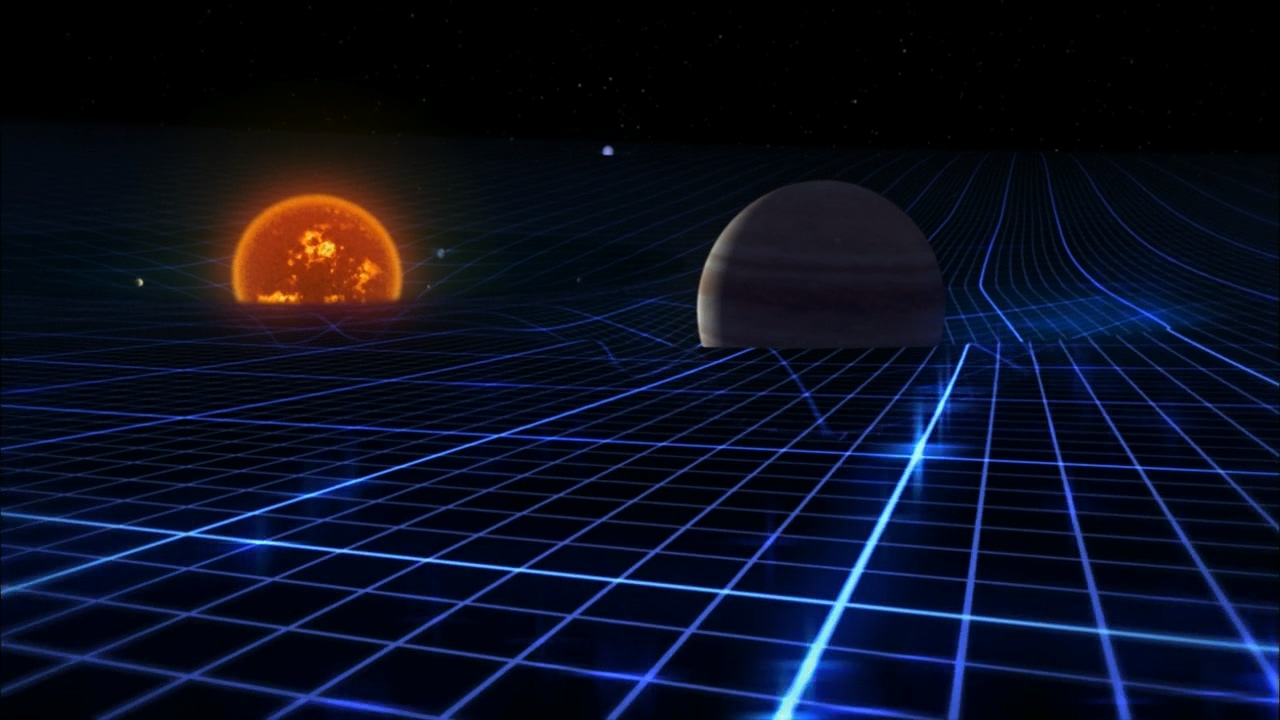} & Curvature occurs in interstellar space. (Relativity) & $\times$ ($\checkmark$) & $\times$ ($\checkmark$) & $\times$ ($\checkmark$) & $\times$ ($\times$) \\
\includegraphics[height=0.82cm,align=c,keepaspectratio]{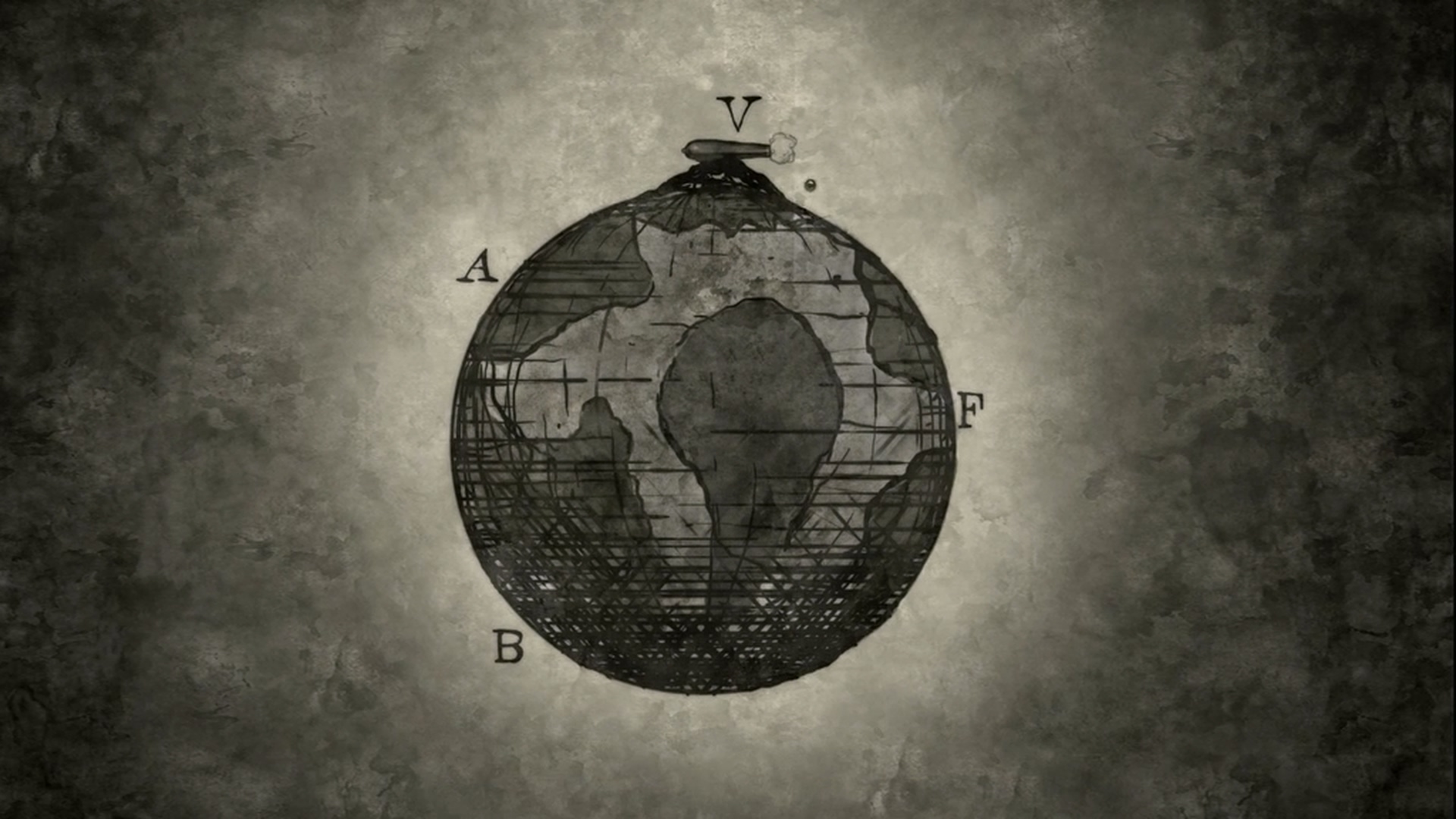} & A cannonball is pulled back to the ground by gravity. (Universal Gravity) & $\checkmark$ ($\checkmark$) & $\checkmark$ ($\checkmark$) & $\checkmark$ ($\checkmark$) & $\times$ ($\checkmark$) \\
\includegraphics[height=0.82cm,align=c,keepaspectratio]{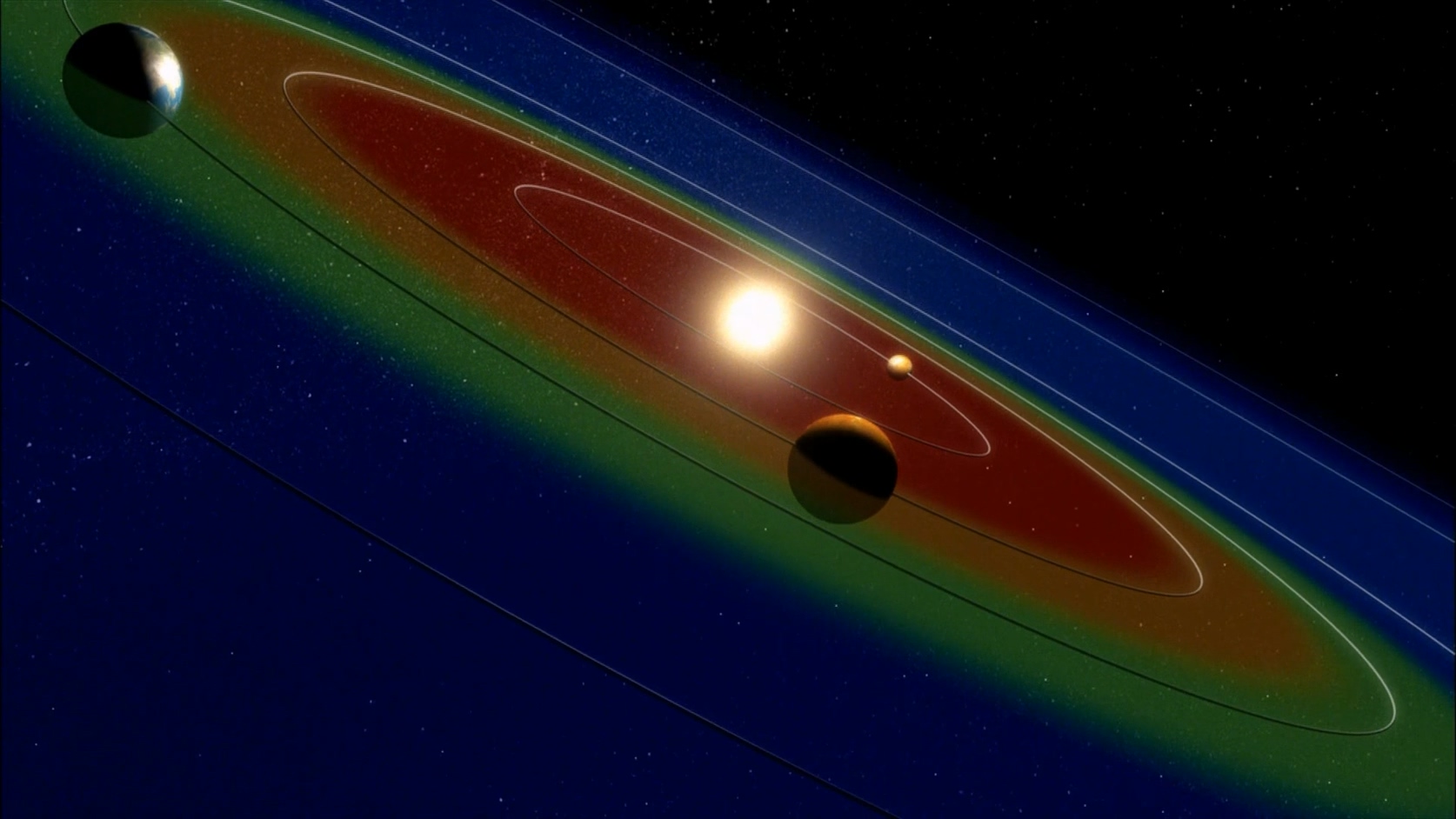} & Planets orbit the Sun in elliptical orbits. (Universal Gravity) & $\checkmark$ ($\times$) & $\checkmark$ ($\checkmark$) & $\checkmark$ ($\times$) & $\checkmark$ ($\checkmark$) \\
\bottomrule
\end{NiceTabular}
\end{adjustbox}
\begin{tablenotes}
\scriptsize
\item The first mark evaluates concrete image--description matching; the bracketed mark evaluates abstract image--concept matching.
\end{tablenotes}
\end{threeparttable}
\end{table*}

\paragraph{Role of textual concept structuring}
The ablations in Tables~\ref{tb:dig_res} and~\ref{tab:ranking_metrics} isolate the source of improvement. Removing Stage~1 reduces MAP@10 from 0.3848 to 0.2974, NDCG@10 from 0.5402 to 0.4309, and the pedagogical score from 24 to 20. Replacing concept labels with literal descriptions performs worse: CDDA(II\&III)$^{d}$ reaches 12 pedagogical points, and appending concept words to descriptions, CDDA(II\&III)$^{d+c}$, reaches only 13. The gain therefore comes from structuring the text space with example--concept supervision before cross-modal adaptation, rather than from exposing the model to concept tokens. Text-only baselines remain weak, suggesting that subtitles alone miss visually expressed teaching evidence.

\begin{figure}[t]
\centering
\scriptsize
\captionsetup[subfigure]{font=scriptsize,skip=1pt}
\captionsetup[figure]{font=small,skip=3pt}
\begin{subfigure}[t]{0.445\textwidth}
\centering
\includegraphics[width=\linewidth]{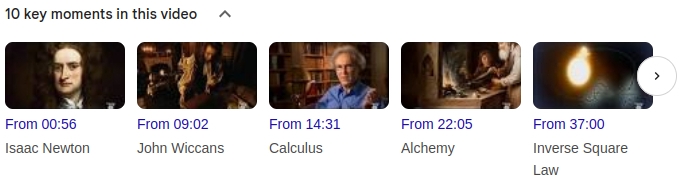}
\caption{Google's 10 key moments (left)}
\end{subfigure}
\hspace{6pt}
\begin{subfigure}[t]{0.445\textwidth}
\centering
\includegraphics[width=\linewidth]{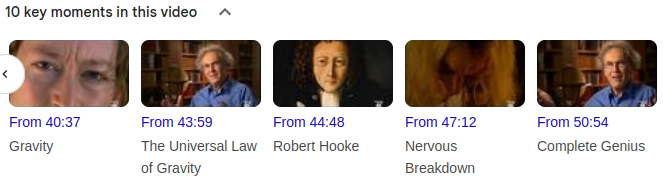}
\caption{Google's 10 key moments (right)}
\end{subfigure}
\vspace{2pt}
\begin{subfigure}[t]{0.445\textwidth}
\centering
\includegraphics[width=\linewidth]{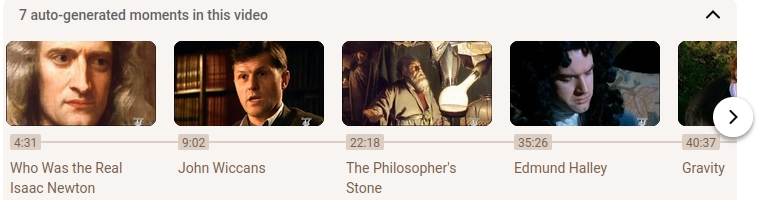}
\caption{YouTube's 7 auto-generated moments (left)}
\end{subfigure}
\hspace{6pt}
\begin{subfigure}[t]{0.445\textwidth}
\centering
\includegraphics[width=\linewidth]{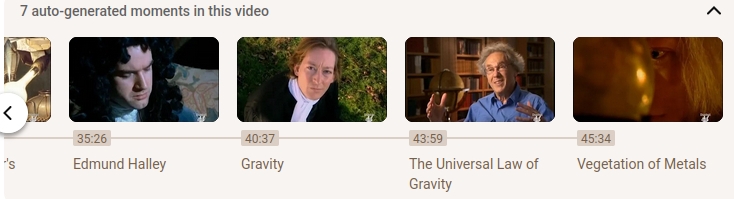}
\caption{YouTube's 7 auto-generated moments (right)}
\end{subfigure}
\vspace{-4pt}
\caption{Commercially highlighted moments for Newton's Dark Secrets collected in April 2026. These screenshots are used only as qualitative context, not as a reproducible baseline.}
\label{fig:commercial_moments}
\end{figure}

\begin{figure}[t]
\centering
\scriptsize
\captionsetup[subfigure]{font=scriptsize,skip=1pt}
\captionsetup[figure]{font=small,skip=3pt}
\begin{subfigure}[t]{0.155\textwidth}
\centering
\includegraphics[width=\linewidth]{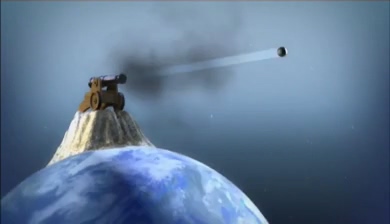}
\caption{if no gravity}
\end{subfigure}
\hspace{2pt}
\begin{subfigure}[t]{0.155\textwidth}
\centering
\includegraphics[width=\linewidth]{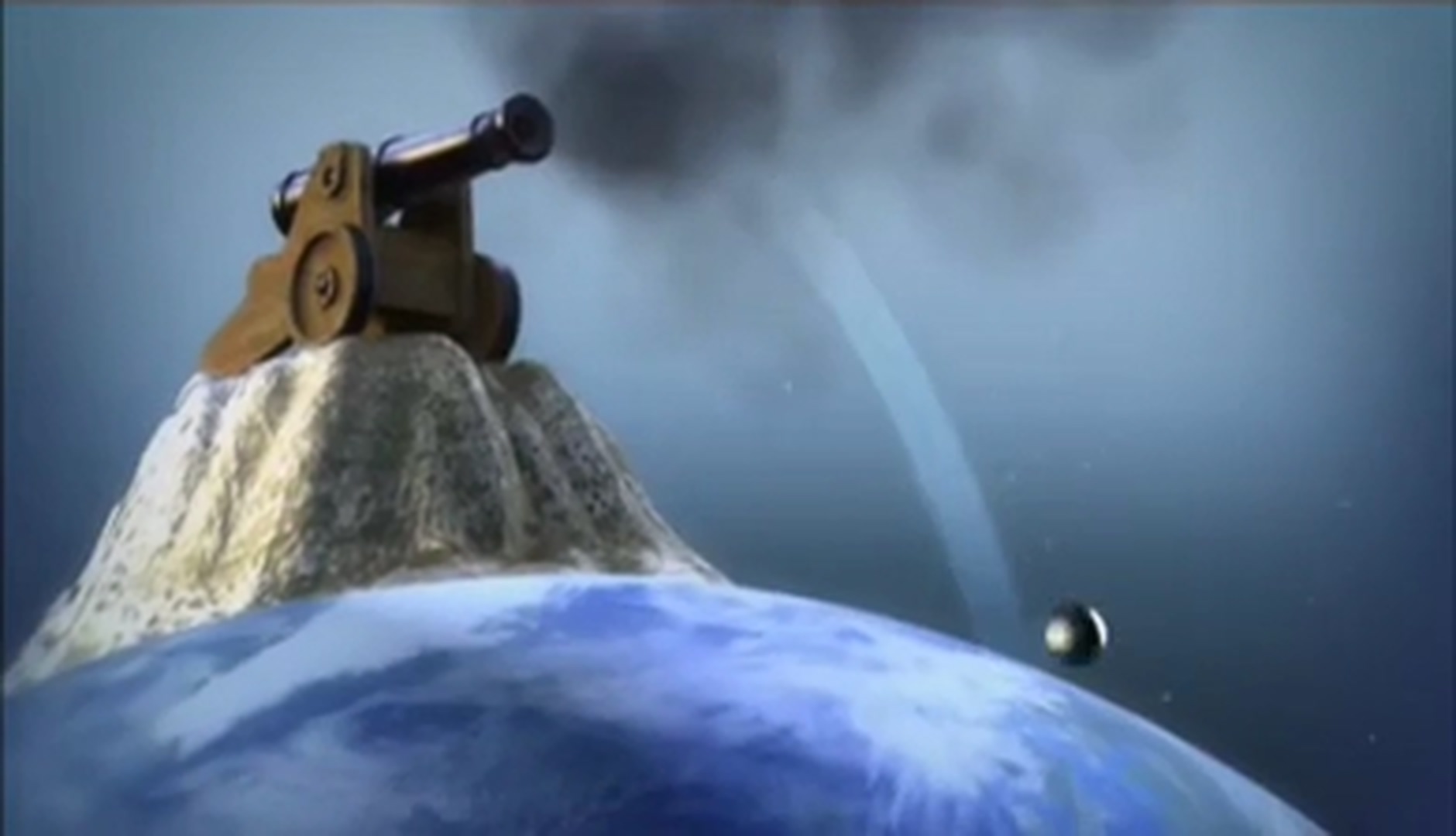}
\caption{low speed}
\end{subfigure}
\hspace{2pt}
\begin{subfigure}[t]{0.155\textwidth}
\centering
\includegraphics[width=\linewidth]{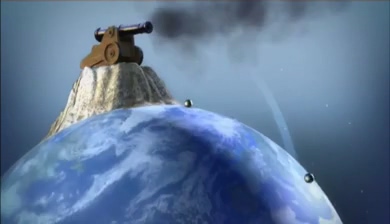}
\caption{different speeds}
\end{subfigure}
\hspace{2pt}
\begin{subfigure}[t]{0.155\textwidth}
\centering
\includegraphics[width=\linewidth]{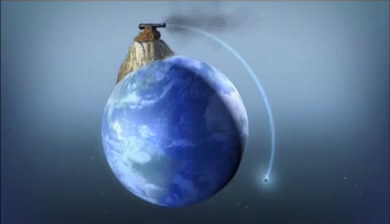}
\caption{fast enough}
\end{subfigure}
\hspace{2pt}
\begin{subfigure}[t]{0.155\textwidth}
\centering
\includegraphics[width=\linewidth]{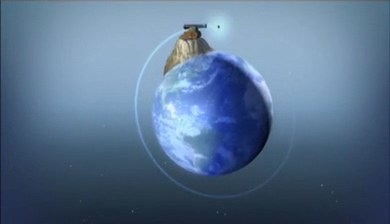}
\caption{circling}
\end{subfigure}
\vspace{-4pt}
\caption{Original Newton's cannonball thought experiment.}
\label{fig:cannonball_original}
\end{figure}

\begin{figure}[!htbp]
\centering
\scriptsize
\captionsetup[subfigure]{font=scriptsize,skip=1pt}
\captionsetup[figure]{font=small,skip=3pt}
\begin{subfigure}[t]{0.13\textwidth}
\centering
\includegraphics[width=\linewidth]{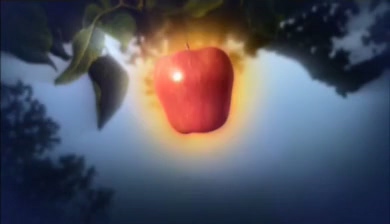}
\caption{An apple}
\end{subfigure}
\hspace{2pt}
\begin{subfigure}[t]{0.13\textwidth}
\centering
\includegraphics[width=\linewidth]{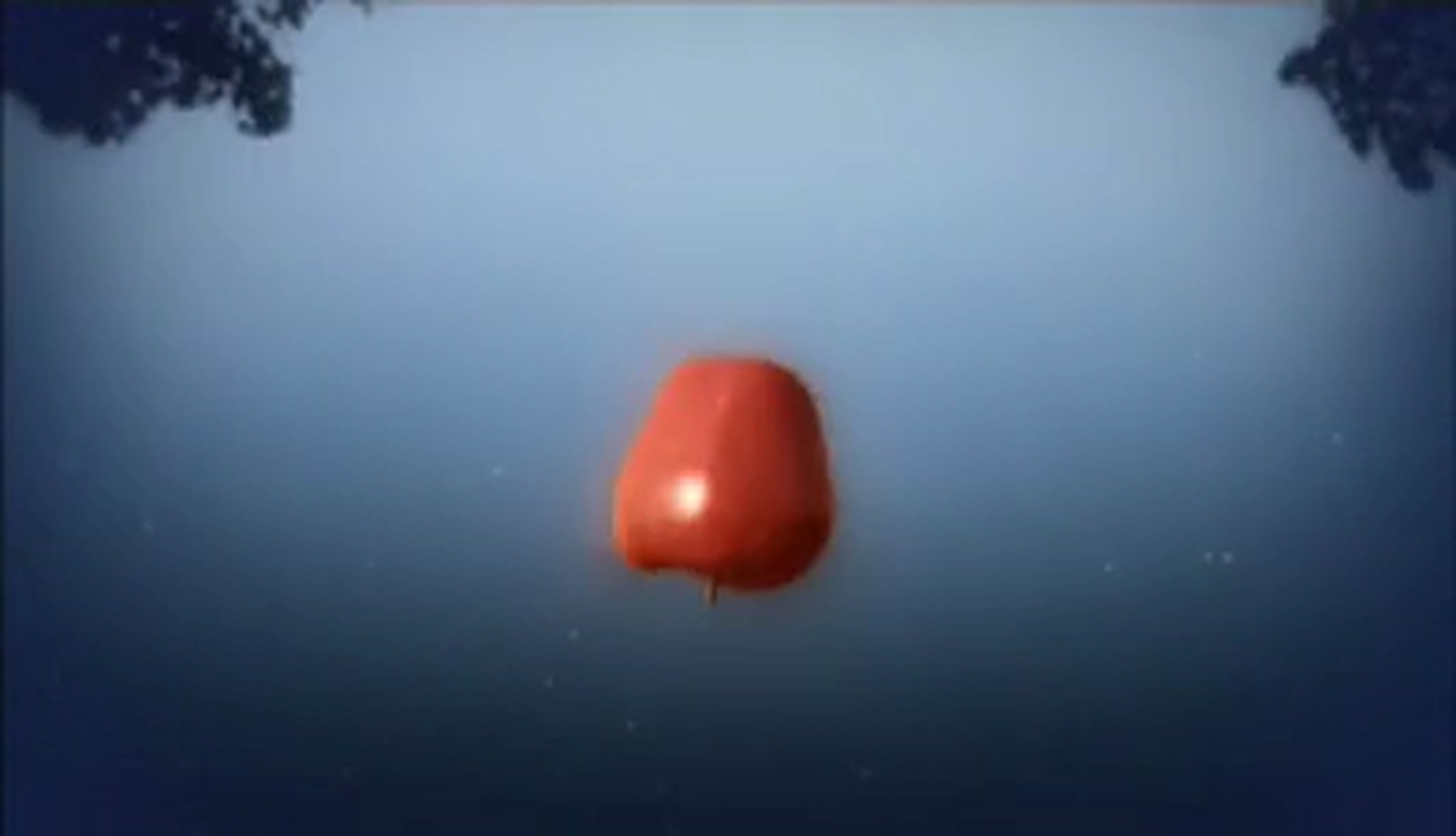}
\caption{falls}
\end{subfigure}
\hspace{2pt}
\begin{subfigure}[t]{0.13\textwidth}
\centering
\includegraphics[width=\linewidth]{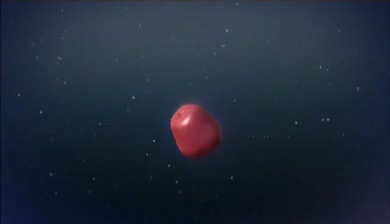}
\caption{into space}
\end{subfigure}
\hspace{2pt}
\begin{subfigure}[t]{0.13\textwidth}
\centering
\includegraphics[width=\linewidth]{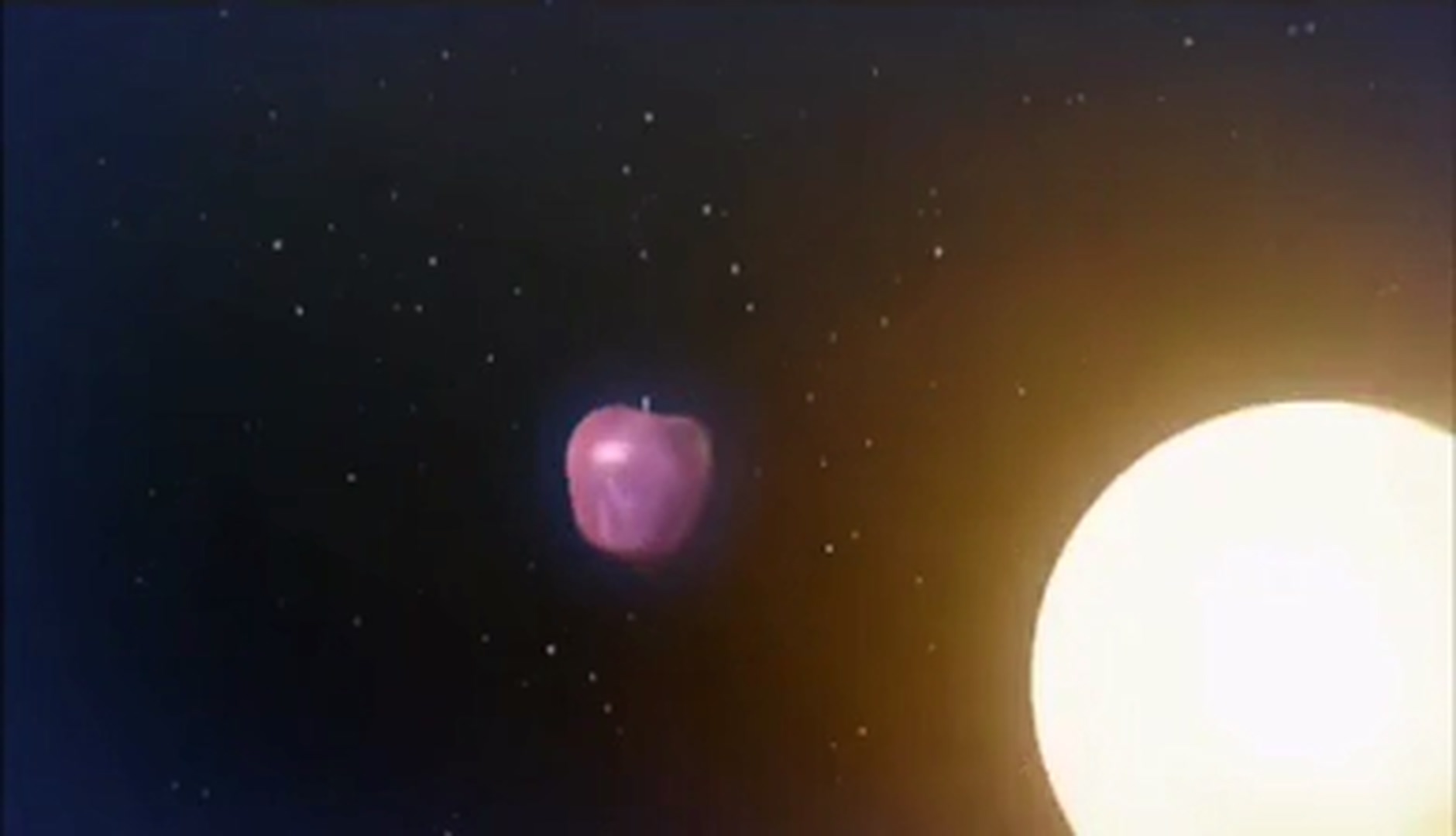}
\caption{becoming}
\end{subfigure}
\hspace{2pt}
\begin{subfigure}[t]{0.13\textwidth}
\centering
\includegraphics[width=\linewidth]{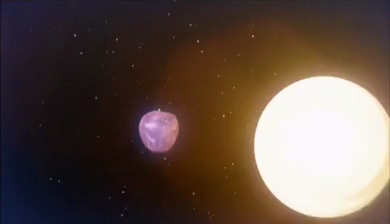}
\caption{the Earth}
\end{subfigure}
\hspace{2pt}
\begin{subfigure}[t]{0.13\textwidth}
\centering
\includegraphics[width=\linewidth]{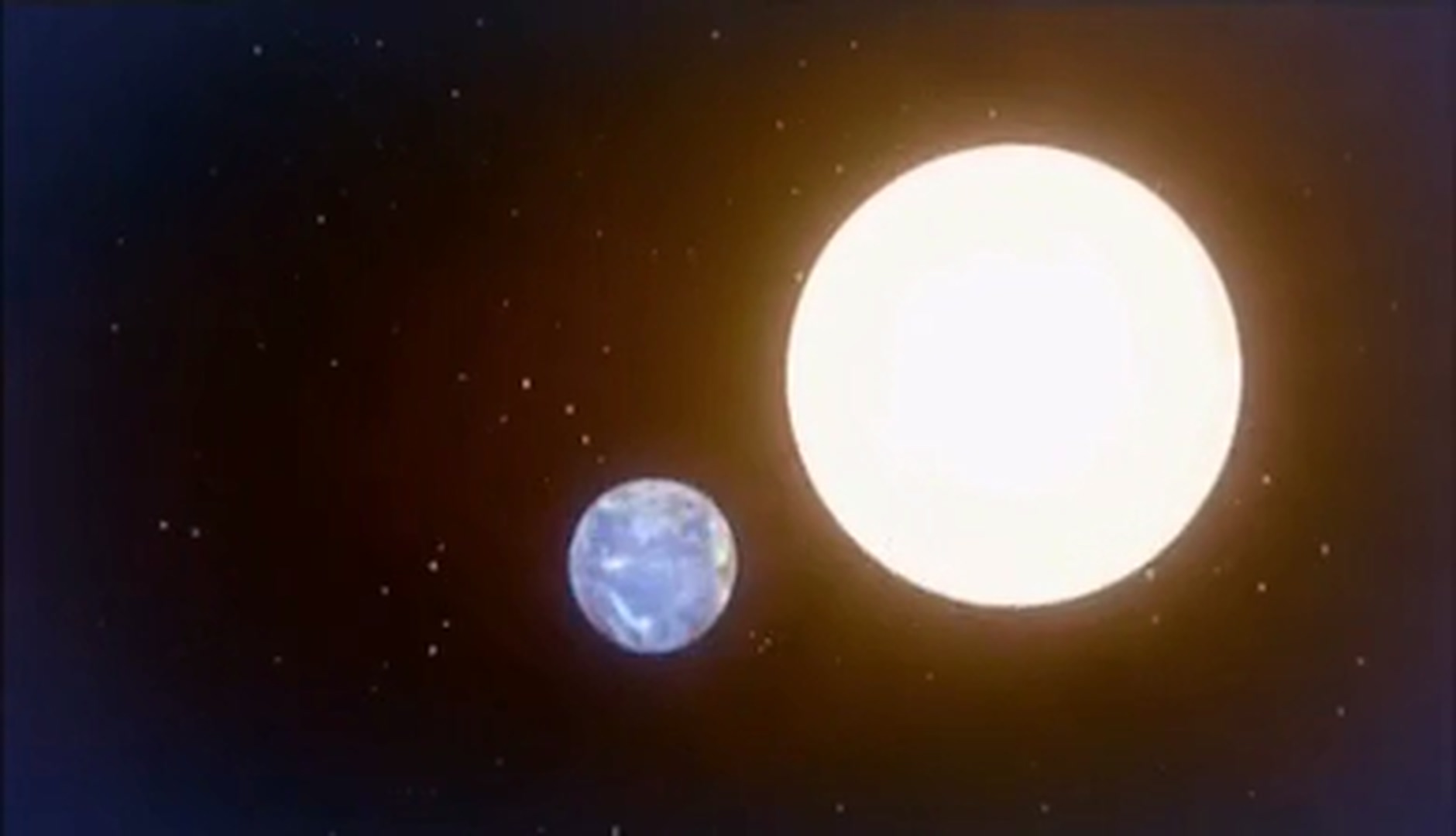}
\caption{around Sun}
\end{subfigure}
\vspace{-4pt}
\caption{Non-trivial match (NtM): a variant of Newton's cannonball thought experiment that bridges Gravity and Universal Gravity.}
\label{fig:ntm_variant}
\end{figure}

\paragraph{Qualitative behavior.}
Fig.~\ref{fig:commercial_moments} shows that generic video highlights are not a reliable substitute for concept-to-example retrieval: they emphasize navigation-friendly biography or person-centered moments, while pedagogically useful evidence may be visually brief and conceptually specific. Figs.~\ref{fig:cannonball_original} and~\ref{fig:ntm_variant} explain the NtM criterion. The apple-to-Earth sequence is broadly related to gravity, but it is most valuable for teaching universal gravitation because it connects near-surface falling with orbital motion. CDDA ranks it under Universal Gravity rather than treating it as generic gravity, which is desirable for curriculum sequencing.

\section{Limitations and Conclusion}

This work is limited by the difficulty of aligning abstract scientific concepts with visually grounded evidence in documentary videos. As discussed in “Newton’s First Law”, one segment may correspond to multiple concepts, making retrieval and evaluation less exact. This issue is further affected by our subtitle-based segmentation, which may produce redundant or incomplete visual evidence when scenes continue without subtitles. Additional analyses of retrieval diversity, pedagogical exactness, model scale, and dataset construction are provided in Appendix~\ref{app:discussion} and Appendix~\ref{app:exp_details}. 
 Future work should expand the benchmark across domains, documentaries, annotators, and resampled splits, and further disentangle visual, subtitle, narration, and audio cues.

We introduced concept-to-example video retrieval, where teachers search long-form documentaries with compact curriculum concepts rather than literal scene descriptions. CDDA bridges this abstraction gap by structuring the text space with educational example--concept pairs and transferring the resulting concept-aware geometry to documentary visuals through staged adaptation. Experiments on our curated physics benchmark show that CDDA improves pedagogically oriented retrieval for abstract and visually diverse concepts while preserving concrete image--description alignment, suggesting that curriculum concepts can serve as effective semantic anchors for instructional video retrieval.

\begin{ack}
We thank Mr. Jiehu Li for his modifications on KBERT's source codes, and we also thank Mr. Xu Wang for his documentary collecting though these documentaries cannot be directly exploited by the Physics curriculum, during their Master Degree Programs in Central China Normal University.

We also thank four anonymous reviewers for their insightful comments on our work.

This version differs slightly from the OpenReview submission. Aside from minor formatting changes and an added explanation
in Subsection 3.7 of why freezing precedes joint training, the content is unchanged. Additional experiments appear only in the
OpenReview rebuttals.
\end{ack}

\FloatBarrier

\begin{small}
\bibliographystyle{IEEEtran}
\bibliography{ref}
\end{small}


\appendix

\section*{Technical Appendices and Supplementary Material}
\addcontentsline{toc}{section}{Technical Appendices and Supplementary Material}
\setcounter{equation}{0}
\renewcommand{\theequation}{A\arabic{equation}}
\setcounter{table}{0}
\setcounter{figure}{0}
\renewcommand{\thetable}{A\arabic{table}}
\renewcommand{\thefigure}{A\arabic{figure}}

\section{Discussion}
\label{app:discussion}
In this section, the diversity of retrieval results, the pedagogical exactness of retrieval results, and the parameter scales are discussed.
\subsection{On the diversity of retrieval results}
\begin{figure}[htbp]
    \centering
    \begin{subfigure}[b]{0.18\textwidth}
        \centering
        \includegraphics[width=\linewidth]{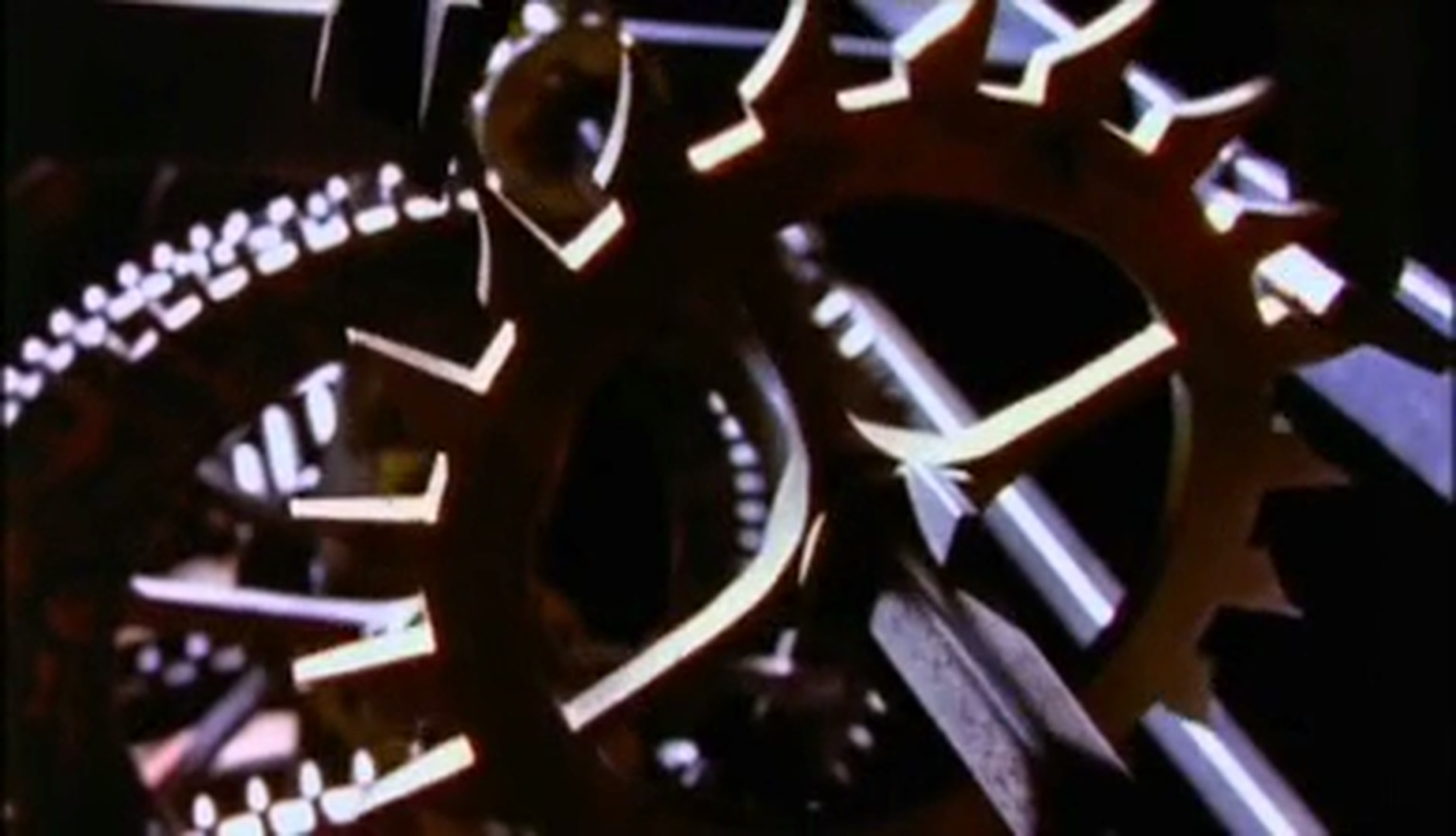}
        \caption{Gear}
        \label{fig:gear}
    \end{subfigure}
    \begin{subfigure}[b]{0.18\textwidth}
        \centering
        \includegraphics[width=\linewidth]{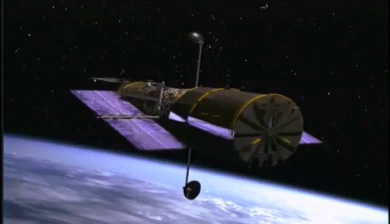}
        \caption{Lever}
        \label{fig:lever}
    \end{subfigure}
    \begin{subfigure}[b]{0.18\textwidth}
        \centering
        \includegraphics[width=\linewidth]{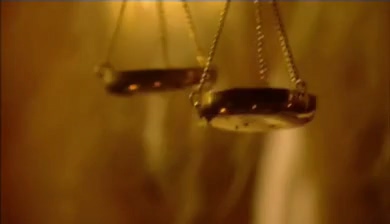}
        \caption{Balance scale}
        \label{fig:balance}
    \end{subfigure}
    \begin{subfigure}[b]{0.18\textwidth}
        \centering
        \includegraphics[width=\linewidth]{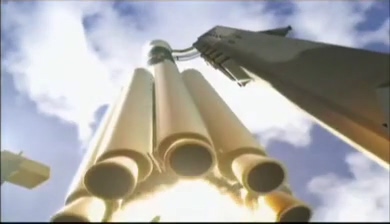}
        \caption{Thrust}
        \label{fig:thrust}
    \end{subfigure}
    \begin{subfigure}[b]{0.18\textwidth}
        \centering
        \includegraphics[width=\linewidth]{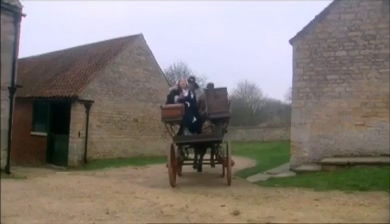}
        \caption{Pulling a cart}
        \label{fig:horse}
    \end{subfigure}
    \caption{Diverse forces in Newton's Dark Secrets besides gravity or universal gravity}
    \label{fig:force}
\end{figure}

Fig. \ref{fig:force} is a collection of the representative images appearing in different segments for the concept of ``Force''. They are all classical mechanical devices in middle school physics textbooks. However, only Fig. \ref{fig:thrust} was hit by the cutting-edge multimodal embedding models (Gemini Embedding 2, Qwen3-VL-Embedding-2B, ImageBind, and InternVideo-Next).

It is noted that this 52.5-minute-long science documentary was divided into 146 segments. These segments are averagely 21.6 seconds long, which is very similar to the common length of short videos on Ticktok or Youtube. Most of these segments exhibit 1 or 2 short but complete scenes. Therefore, it is very unlikely that these cutting-edge multimodal embedding models do not understand what a specific scene is, but without the concept driven idea, they just fail to relate the current scene to a superordinate curricular concept. Or it seems that they do not realize that \emph{gears} themselves can be used to exhibit a form of force in a middle school physics class, though in this documentary it is used as a metaphor to the gear-driven precision of the universe.

Similar situations happen in the scenes when this documentary is exhibiting Hubble Space Telescope (Fig. \ref{fig:lever}) equipped with a \emph{lever}-like solar panel (new material for the ``Force'' curricular unit), restaging Newton's obsession with alchemy when the actor holding a \emph{balance scale}, and telling the audience Newton sitting in a cart \emph{pulled} by a horse to flee from the plague. What these embedding models returned when ``Force'' is the query are largely overlapped with the retrieval results of the query ``Gravity'' or ``Universal Gravity''.

\subsection{On the pedagogical exactness of retrieval results}
Given the high pedagogical value of the non-trivial segment shown in Fig. \ref{fig:ntm_variant}, the majority of following discussion will revolve around it.

According to Table \ref{tb:dig_res}, the original CN-CLIP can recognize the non-trivial segment when the query is either ``Gravity'' or ``Universal Gravity''. As aforementioned in subsection  \ref{subsec:main_results}, this non-trivial segment is not a quite convincingly exact match for the concept \emph{gravity} in pedagogy, particularly when teachers are about to introduce this concept. In terms of the structure and training paradigm (2-stage training and training with description-image pairs \emph{not} concept-image pairs), CDDA (II\&III)\textsuperscript{d} is a close relative of CN-CLIP, in contrast to CDDA. Table \ref{tb:dig_res} indicates that CDDA (II\&III)\textsuperscript{d} inherits the pedagogical inexact match on this non-trivial segment.

On the other hand, due to a 3-stage training with ``Gravity''-example pairs (including in language and visual modals), CDDA successfully avoids returning this non-trivial segment as a top 10 retrieval result for the ``Gravity'' query. As its close relative, CDDA (II\&III) which is also trained with concept-image pairs, also succeeds on this matter.

By contrasting the inexactness with the exactness of this NtM, the effectiveness and value of training with concept-example pairs are demonstrated. What's more, CDDA (II\&III)\textsuperscript{d+c}'s performance on this non-trivial segment, as shown in Table \ref{tb:dig_res}, confirms again that Stage 1's training is necessary: simply squeezing the description of a phenomenon and its conveyed concept into one single text item in a text-image pair is not the most effective.

Speaking of ``Newton's First Law'', we did not have any visual examples before this experiment until we found Fig. \ref{fig:ntm_variant} in this ``Newton's Dark Secrets'' documentary after the experiment, because this concept is so abstract that it can only exist in an extremely ideal situation. Nevertheless, if we simply use a cannonball flying picture as the example part in the ``Newton's First Law''-example pair, we are afraid that it may undermine the embedding space of CDDA because its base, CN-CLIP, may already have adequate reasons or sufficient prior knowledge to ``mistake'' this cannonball flying picture as an example of shooting or firing. So we ruled this kind of pictures out of our training dataset. Instead, we paired a picture similar to Fig. \ref{fig:newtonfirstlawinput} with the concept ``Newton's First Law'' for training, and we labeled the segment which shows the page containing Newton's First Law in his monograph, as the ground truth for the concept query. However, no matter ``almost'' hit or hit, such segment is less inspiring in a middle school physics class than the one that contains Fig. \ref{fig:cannonball_original}. Gemini Embed. 2 is the only model in Table \ref{tb:dig_res} that returns the segment which contains Fig. \ref{fig:cannonball_original}, and ranks it at 5th in its top 10 retrieval results for the ``Newton's First Law'' query.

To sum up, ``Force'' examines the diversity of retrieval results, ``Gravity'' and ``Universal Gravity'' examine the pedagogical exactness of retrieval results, and ``Newton's First Law'' shows the ceiling-level difficulty of pedagogical exactness. Gemini Embed 2 shows its outstanding performance on the last one. However, it fails at the ``Newton'' query.
\subsection{On the Gemini Embed. 2 's retrieval results for the ``Newton'' query}
The ground truth of the retrieval results for the ``Newton'' query are those segments which contain Newton's portrait. It was originally designed as a benchmark. However, it seems that Gemini Embed. 2 ``intentionally'' avoids any segment that contains Newton's portrait. The top 10 retrieval results returned by Gemini Embed. 2 are all segments that either contain actors who play the role of Newton, or contain narratives on Newton's contributions. These contributions were either restaged by an actor or commented by a professional in this documentary.

At first, one may misunderstand that Gemini Embed. 2 does not recognize Newton. In the design philosophy of Gemini Embed. 2, the core essence of Newton this symbol is probably his contributions, which is a reasonable speculation from \cite{Kusupati2022}. If such design philosophy can be rephrased into a style similar to ``concept-example'', ``symbol-essence'' may probably be the best expression. This difference is reflected into the understanding on the intention behind the query ``Newton''. However, this ``symbol-essence'' style may still remain at the ``term-description'' semantical level, where the ``term'' and ``description'' have the same semantics as in a dictionary.
\subsection{On the parameter scale}
The parameter scales of the five large models are presented in Table \ref{tb:parameterscales}. As seen, CDDA adopts a relatively small model among these models, CN-CLIP, but achieves competitive results. However, CN-CLIP can only obtain a video's embedding by sampling the video's frames and pooling their embeddings. To be fair enough, Qwen3-VL-Embedding, ImageBind and InternVideo-Next have to follow the same way though they can treat a video segment as a whole. Gemini-Embedding-2 is the only model that directly returns a video segment's embedding due to the consideration of saving remote calling its APIs.
\begin{table}[!hbtp]
  \caption{Parameter scales of models}
  \label{tb:parameterscales}
  \centering
  \begin{threeparttable}
  \begin{tabular}{lrr}
    \toprule
    Model Name and Version          &  Parameter Scales  & Checkpoint File Size\\
    \midrule
    Gemini-Embedding-2              &  Not Officially Declared & Not Officially Declared\\
    Qwen3-VL-Embedding-2B           &  2B                      & 4.3  GB\\
    ImageBind-huge                  &  Not Officially Declared & 4.8  GB \\
    InternVideo-Next\textsubscript{large\_p14\_res224\_f16)} & 0.3B \tnote{1} & 622.1 MB\\
    CN-CLIP\textsubscript{ViT-B/16} & 188M \cite{Yang2023}  & 753.2 MB\\
    \bottomrule
  \end{tabular}
  \begin{tablenotes}
  \item [1] \small{\url{https://huggingface.co/revliter/internvideo_next_large_p14_res224_f16}}
  \end{tablenotes}
  \end{threeparttable}
\end{table}

\section{Supplementary Proofs and Concept-Alignment Diagnostics}

\subsection{Proof of Eq. (7) in Subsection \ref{the:consistrans}}
\label{subsectiontheproof}
\begin{proof}
According to their definitions in Eq. (7) in Subsection \ref{the:consistrans}, the vectors satisfy the following equations:
\begin{equation}
\begin{aligned}
\mathbf{v} &= \mathbf{v}_{\parallel \mathbf{c}} + \mathbf{v}_{\perp \mathbf{c}} \\
\mathbf{u} &= \mathbf{u}_{\parallel \mathbf{c}} + \mathbf{u}_{\perp \mathbf{c}}
\end{aligned}
\end{equation}

Then,
\begin{equation}
\begin{aligned}
\mathbf{v} \cdot \mathbf{u} &= (\mathbf{v}_{\parallel \mathbf{c}} + \mathbf{v}_{\perp \mathbf{c}}) \cdot (\mathbf{u}_{\parallel \mathbf{c}} + \mathbf{u}_{\perp \mathbf{c}}) \\
                            &= \mathbf{v}_{\parallel \mathbf{c}} \cdot \mathbf{u}_{\parallel \mathbf{c}} + \mathbf{v}_{\perp \mathbf{c}} \cdot \mathbf{u}_{\perp \mathbf{c}}
\end{aligned}
\end{equation}

\begin{equation}
\begin{aligned}
\mathbf{v}_{\parallel \mathbf{c}} \cdot \mathbf{u}_{\parallel \mathbf{c}} &= \|\mathbf{v}\| \cos \theta_{\mathbf{v}\mathbf{c}}\|\mathbf{u}\| \cos \theta_{\mathbf{u}\mathbf{c}} \\
                                                                   &= \|\mathbf{v}\| \|\mathbf{u}\| \cos \theta_{\mathbf{v}\mathbf{c}} \cos \theta_{\mathbf{u}\mathbf{c}}
\end{aligned}
\end{equation}

Since $\|\mathbf{v}_{\perp \mathbf{c}}\| = \|\mathbf{v}\|\sin \theta_{\mathbf{v}\mathbf{c}}$ and $\|\mathbf{u}_{\perp \mathbf{c}}\| = \|\mathbf{u}\|\sin \theta_{\mathbf{u}\mathbf{c}}$,
\begin{equation}
\begin{aligned}
\mathbf{v}_{\perp \mathbf{c}} \cdot \mathbf{u}_{\perp \mathbf{c}} &= \|\mathbf{v}_{\perp \mathbf{c}}\|\|\mathbf{u}_{\perp \mathbf{c}}\| \cos \phi \\
                                                           &= \|\mathbf{v}\| \sin \theta_{\mathbf{v}\mathbf{c}} \|\mathbf{u}\| \sin \theta_{\mathbf{u}\mathbf{c}} \cos \phi
\end{aligned}
\end{equation}

$\phi$ is also defined in the context of Eq. (7) in Subsection \ref{the:consistrans}.

Therefore,
\begin{equation}
\begin{aligned}
\cos \theta_{\mathbf{v}\mathbf{u}} &= \frac{\mathbf{v} \cdot \mathbf{u}}{\|\mathbf{v}\|\|\mathbf{u}\|}
&= \cos \theta_{\mathbf{v}\mathbf{c}}\cos \theta_{\mathbf{u}\mathbf{c}}+\sin \theta_{\mathbf{v}\mathbf{c}}\sin \theta_{\mathbf{u}\mathbf{c}}\cos \phi
\end{aligned}
\end{equation}

By (A5), we can conclude that if $\theta_{\mathbf{v}\mathbf{c}}$ and $\theta_{\mathbf{u}\mathbf{c}}$ can be both trained to be small, say they are both between 0 to 90 degrees, $\cos \theta_{\mathbf{v}\mathbf{u}}$ will still be positive as long as $\phi$ is also between 0 to 90 degrees. When $\phi$ is between 90-180 degrees, $\cos \theta_{\mathbf{v}\mathbf{u}}$ still has very large chance to be positive by (A5). In fact, a simple trigonometric function calculation suffices to prove that if $\theta_{\mathbf{v}\mathbf{c}}$ and $\theta_{\mathbf{u}\mathbf{c}}$ are both between 0 to 45 degrees, no matter how large $\phi$ is, $\cos \theta_{\mathbf{v}\mathbf{u}}$ will be positive. This positivity indicates that $\theta_{\mathbf{v}\mathbf{u}}$ is between 0 to 90 degrees, where $\mathbf{v}$ and $\mathbf{u}$ can be considered semantically similar to some extent.
\end{proof}

\subsection{Samples of Concept--Example Pairs}
\label{subsectionconcept_examplesamples}

Stage~1 uses 150 textual concept--example pairs to structure the concept-aware text space before cross-modal adaptation. These pairs cover six curriculum or auxiliary concepts: 35 pairs for Force, 28 for Gravity, 14 for Newton, 18 for Newton's First Law, 51 for Universal Gravity, and 4 for Relativity. The pairs are split into 103 training, 31 validation, and 16 test items. Table~\ref{tab:concept_example_samples} gives representative samples. The examples are not intended to be literal captions; instead, each item describes a phenomenon, fact, or explanation that can instantiate the corresponding concept in instruction.

\begin{table}[!htbp]
\centering
\small
\caption{Representative textual concept--example pairs used in Stage~1 of CDDA.}
\label{tab:concept_example_samples}
\begin{tabular}{p{0.23\linewidth}p{0.68\linewidth}}
\toprule
Concept & Example / fact \\
\midrule
\multirow{2}{*}{Force}
& The tires, pedals, or handlebars of a bicycle are engraved with uneven patterns to increase friction by increasing the roughness of the contact surface. \\
& A person kicks a ball; a horse pulls a cart; a person pushes a table. \\
\midrule
\multirow{2}{*}{Gravity}
& The effect of gravity on an object in free fall is to make its velocity increase. \\
& An astronaut experiences weightlessness in the Space Station. \\
\midrule
\multirow{2}{*}{Newton}
& In 1664, force was defined as the time derivative of momentum. \\
& Newton's three laws of motion were formulated as axioms. \\
\midrule
\multirow{2}{*}{Newton's First Law}
& Due to inertia, a bullet continues to fly forward at the speed it had when it left the gun barrel; no external force is required to maintain this motion. \\
& If no air resistance or gravity acted on the bullet, it would keep moving forward uniformly and indefinitely. \\
\midrule
\multirow{2}{*}{Universal Gravity}
& Newton's cannonball thought experiment shows that Earth's attraction to a cannonball and Earth's attraction to the Moon can be understood as the same type of force. \\
& Jupiter's moons orbit Jupiter in elliptical paths. \\
\bottomrule
\end{tabular}
\end{table}

\subsection{Details on the Mapping Preservation Examination}
\label{subsectionpreservation}

Table~\ref{tb:details-preservation} provides the detailed outputs behind the mapping-preservation diagnostic in Table~\ref{tab:preservation}. This diagnostic checks whether concept-driven adaptation preserves concrete image--description matching while improving image--concept assignment. The rows compare the ground-truth description and concept with the outputs produced by CN-CLIP, CDDA, CDDA(I\&II), and Qwen3-VL-2B. Phrases such as ``identical to CN-CLIP'' refer to the generated description only; the concept in parentheses reports the corresponding concept prediction.

\begin{table}[!htbp]
\centering
\scriptsize
\caption{Detailed outputs for the mapping-preservation examination. The text before parentheses is the image description; the term in parentheses is the associated or predicted concept.}
\label{tb:details-preservation}
\begin{adjustbox}{max width=\linewidth,center}
\begin{tabular}{c p{0.55\linewidth} l}
\toprule
Test image & Output description and concept & Producer \\
\midrule
\multirow{5}{*}{\includegraphics[width=2.4cm]{59.jpg}}
& An astronaut bounce-walks on the Moon. (Gravity) & Ground truth \\
& An astronaut bounce-walks on the Moon. (Universal Gravity) & CN-CLIP \\
& Identical to CN-CLIP. (Gravity) & CDDA \\
& Identical to CN-CLIP. (Gravity) & CDDA(I\&II) \\
& Identical to the ground-truth description. (Universal Gravity) & Qwen3-VL-2B \\
\midrule
\multirow{5}{*}{\includegraphics[width=2.4cm]{31.jpg}}
& An astronaut experiences weightlessness in the Space Station. (Gravity) & Ground truth \\
& An astronaut experiences weightlessness in the Space Station. (Gravity) & CN-CLIP \\
& Identical to CN-CLIP. (Gravity) & CDDA \\
& Identical to CN-CLIP. (Gravity) & CDDA(I\&II) \\
& Identical to the ground-truth description. (Relativity) & Qwen3-VL-2B \\
\midrule
\multirow{5}{*}{\includegraphics[width=2.4cm]{36.jpg}}
& An astronaut experiences weightlessness in a plane. (Gravity) & Ground truth \\
& An astronaut experiences weightlessness in the Space Station. (Gravity) & CN-CLIP \\
& Identical to CN-CLIP. (Gravity) & CDDA \\
& Identical to CN-CLIP. (Gravity) & CDDA(I\&II) \\
& An astronaut experiences weightlessness in the Space Station. (Gravity) & Qwen3-VL-2B \\
\midrule
\multirow{5}{*}{\includegraphics[width=2.4cm]{18.jpg}}
& Curvature occurs in interstellar space. (Relativity) & Ground truth \\
& Planets orbit the Sun in elliptical orbits. (Relativity) & CN-CLIP \\
& Identical to CN-CLIP. (Relativity) & CDDA \\
& Identical to CN-CLIP. (Relativity) & CDDA(I\&II) \\
& Planets orbit the Sun in elliptical orbits. (Universal Gravity) & Qwen3-VL-2B \\
\midrule
\multirow{5}{*}{\includegraphics[width=2.4cm]{25.jpg}}
& A cannonball is pulled back to the ground by gravity. (Universal Gravity) & Ground truth \\
& A cannonball is pulled back to the ground by gravity. (Universal Gravity) & CN-CLIP \\
& Identical to CN-CLIP. (Universal Gravity) & CDDA \\
& Identical to CN-CLIP. (Universal Gravity) & CDDA(I\&II) \\
& Planets orbit the Sun in elliptical orbits. (Universal Gravity) & Qwen3-VL-2B \\
\midrule
\multirow{5}{*}{\includegraphics[width=2.4cm]{51.jpg}}
& Planets orbit the Sun in elliptical orbits. (Universal Gravity) & Ground truth \\
& Planets orbit the Sun in elliptical orbits. (Relativity) & CN-CLIP \\
& Identical to CN-CLIP. (Universal Gravity) & CDDA \\
& Identical to CN-CLIP. (Relativity) & CDDA(I\&II) \\
& Identical to the ground-truth description. (Universal Gravity) & Qwen3-VL-2B \\
\bottomrule
\end{tabular}
\end{adjustbox}
\end{table}

\section{Additional Experimental Details}
\label{app:exp_details}

This section consolidates the additional details needed to reproduce and interpret the experiments in Section~\ref{sec:experiments}. It covers dataset construction, candidate segmentation, baseline protocols, ablation variants, training settings, metric definitions, InternVideo-Next query images, and generated cannonball illustrations. All tables and figures below are tied to the same held-out retrieval corpus used in the main experiments.

\subsection{Dataset Construction and Annotation}
\label{app:dataset_annotation}

Table~\ref{tab:app_data_summary} summarizes the supervision and evaluation data. The image--concept examples are collected from the documentaries listed in Table~\ref{tab:source_material} and from Motion Mountain (Vol.~1). Each image is annotated by the curriculum concept it can instantiate in instruction, rather than by a literal visual caption. The evaluation documentary, Newton's Dark Secrets, is held out from all adaptation stages.

\begin{table}[!htbp]
\centering
\small
\caption{Summary of supervision and evaluation data.}
\label{tab:app_data_summary}
\begin{tabular}{lcc}
\toprule
Data source & Size & Usage \\
\midrule
Image--concept examples & 44 / 12 / 6 & Train / validation / test diagnostic \\
Textual example--concept pairs & 118 + 32 = 150 & Stage~1 text structuring \\
Evaluation documentary subtitles & 832 & Candidate segmentation \\
Evaluation candidate segments & 146 & Retrieval evaluation \\
Concept queries & 5 & Retrieval queries \\
\bottomrule
\end{tabular}
\end{table}

The five retrieval queries are Force, Gravity, Newton, Newton's First Law, and Universal Gravity. Table~\ref{tab:app_relevance_counts} reports the number of binary-relevant segments for each query in the 146-segment evaluation corpus. These counts define the denominators for MAP@K and the ideal rankings used in NDCG@K.

\begin{table}[!htbp]
\centering
\small
\caption{Number of relevant segments per concept query in the 146-segment evaluation corpus.}
\label{tab:app_relevance_counts}
\begin{tabular}{lc}
\toprule
Concept query & Relevant segments \\
\midrule
Force & 6 \\
Gravity & 20 \\
Newton & 9 \\
Newton's First Law & 1 \\
Universal Gravity & 25 \\
\bottomrule
\end{tabular}
\end{table}

Two senior pedagogical experts cross-check the ground-truth labels. A segment is marked relevant when it can serve as a concrete teaching example of the queried concept, even if the concept word does not appear in subtitles. The non-trivial match (NtM) in Fig.~\ref{fig:ntm_variant} is a special pedagogical case: it depicts an apple falling outward and becoming the Earth orbiting the Sun, thereby connecting near-surface gravity to universal gravitation. In the top-10 pedagogical analysis, NtM receives additional pedagogical credit only for the Universal Gravity query; in MAP/NDCG it is treated as a single binary-relevant segment.

\subsection{Candidate Segmentation}
\label{app:candidate_segmentation}

The held-out documentary is segmented once before model comparison. Following the subtitle-embedding procedure of~\cite{Ghinassi2024}, subtitles are embedded, adjacent subtitles are grouped by semantic continuity, and the resulting groups are converted into non-overlapping candidate segments. This produces 146 fixed candidate segments from 832 subtitles. Because the segmentation is fixed before evaluation, all visual, multimodal, and text-only baselines are compared on exactly the same retrieval pool; only the segment representation and query representation vary across methods.

\subsection{Model and Baseline Protocols}
\label{app:baseline_protocols}

CDDA uses CN-CLIP~\cite{Yang2023} as the base two-tower vision--language model and Chinese RoBERTa~\cite{Cui2021} for the text-side adaptation component. For each video segment, frame or sub-clip embeddings are pooled into one segment embedding. Concept queries are encoded as text and ranked by cosine similarity against segment embeddings.

\begin{figure}[t]
    \centering
    \begin{subfigure}[b]{0.18\textwidth}
        \centering
        \includegraphics[width=\linewidth]{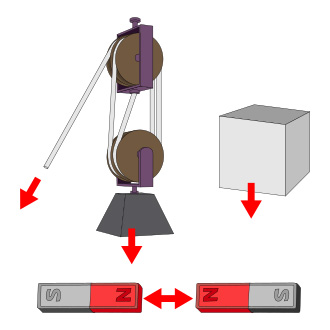}
        \caption{Force}
        \label{fig:forceinput}
    \end{subfigure}
    \hfill
    \begin{subfigure}[b]{0.18\textwidth}
        \centering
        \includegraphics[width=\linewidth]{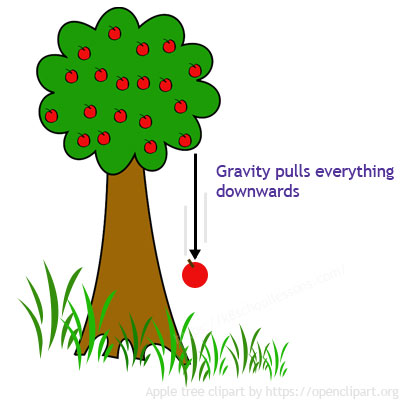}
        \caption{Gravity}
        \label{fig:gravityinput}
    \end{subfigure}
    \hfill
    \begin{subfigure}[b]{0.18\textwidth}
        \centering
        \includegraphics[width=\linewidth]{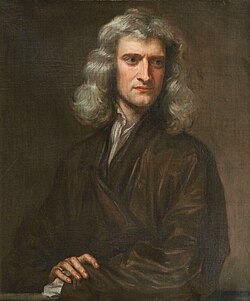}
        \caption{Newton}
        \label{fig:newtoninput}
    \end{subfigure}
    \hfill
    \begin{subfigure}[b]{0.18\textwidth}
        \centering
        \includegraphics[width=\linewidth]{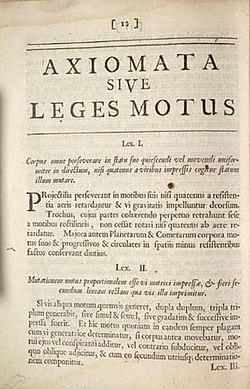}
        \caption{Newton's F. L.}
        \label{fig:newtonfirstlawinput}
    \end{subfigure}
    \hfill
    \begin{subfigure}[b]{0.18\textwidth}
        \centering
        \includegraphics[width=\linewidth]{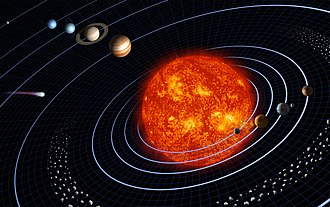}
        \caption{Universal G.}
        \label{fig:universalgravityinput}
    \end{subfigure}
    \caption{Five representative concept images used as InternVideo-Next query inputs. F. L. is short for First Law, and G. is short for Gravity.}
    \label{fig:InternVideoImagesInput}
\end{figure}

For models with a native text--video or text--image retrieval interface, each candidate segment is encoded and ranked by similarity to the concept query. For text-only baselines, retrieval is performed over the subtitle text belonging to each candidate segment. InternVideo-Next~\cite{Wang2026} is used as a visual-retrieval diagnostic rather than as a direct concept-text retriever: each concept is represented by one manually selected query image, and the same 146 candidate segments are ranked under that visual query. Fig.~\ref{fig:InternVideoImagesInput} shows the five query images used for this diagnostic.

Table~\ref{tab:app_ablation_variants} defines the CDDA variants used to isolate the contribution of textual concept structuring and concept-level labels. These variants correspond to the ablation rows reported in Table~\ref{tb:dig_res} and Table~\ref{tab:ranking_metrics}.

\begin{table}[!htbp]
\centering
\small
\caption{Ablation variants used in the retrieval experiments.}
\label{tab:app_ablation_variants}
\begin{tabular}{lp{0.62\linewidth}}
\toprule
Variant & Difference from CDDA \\
\midrule
CDDA(II\&III) & Removes Stage~1 textual concept structuring and trains only with image--concept pairs. \\
CDDA(II\&III)$^{d}$ & Replaces abstract concept labels with literal fact descriptions. \\
CDDA(II\&III)$^{d+c}$ & Appends a concept sentence to each fact description, e.g., ``Planets orbit the Sun in elliptical orbits. This is universal gravity.'' \\
CDDA(II\&III)$^{d}$\textsubscript{text} & Text-only retrieval counterpart of the description-based variant. \\
\bottomrule
\end{tabular}
\end{table}

\subsection{Implementation Details}
\label{app:impl_details}

Table~\ref{tab:app_training_params} lists the training and inference settings used for CDDA in the reported experiments. The visual adaptation stages use the same batch size, learning rate, epoch number, and temperature. The validation split is used for model selection, and the six held-out test images are used only for the preservation diagnostic in Table~\ref{tb:details-preservation}.

\begin{table}[!htbp]
\centering
\small
\caption{Training and inference settings for CDDA.}
\label{tab:app_training_params}
\begin{tabular}{ll}
\toprule
Item & Value \\
\midrule
Base model & CN-CLIP~\cite{Yang2023} \\
Text adaptation component & Chinese RoBERTa~\cite{Cui2021} \\
Optimizer & AdamW \\
Batch size, Stage~2 / Stage~3 & 4 / 4 \\
Learning rate, Stage~2 / Stage~3 & $3\times10^{-6}$ / $3\times10^{-6}$ \\
Epochs, Stage~2 / Stage~3 & 15 / 15 \\
Temperature $\tau_t$ / $\tau_v$ & 0.01 / 0.01 \\
Hardware & NVIDIA GeForce RTX 3090 GPU, 24 GB memory \\
\bottomrule
\end{tabular}
\end{table}

\subsection{Metric Definitions}
\label{app:metric_definitions}

The main ranking metrics are computed with binary relevance. Let $R_q$ be the set of relevant segments for query $q$, and let $\mathrm{rel}_q(k)\in\{0,1\}$ indicate whether the segment ranked at position $k$ is relevant. Precision at rank $k$ is defined as
\begin{equation}
P_q(k)=\frac{1}{k}\sum_{j=1}^{k}\mathrm{rel}_q(j).
\end{equation}
Average precision at cutoff $K$ is computed as
\begin{equation}
\mathrm{AP@}K(q)=\frac{1}{\min(K,|R_q|)}\sum_{k=1}^{K} P_q(k)\,\mathrm{rel}_q(k).
\end{equation}
MAP@K is the mean over the five concept queries:
\begin{equation}
\mathrm{MAP@}K=\frac{1}{|\mathcal{Q}|}\sum_{q\in\mathcal{Q}}\mathrm{AP@}K(q),
\end{equation}
where $\mathcal{Q}$ is the set of concept queries.

For NDCG, discounted cumulative gain is computed as
\begin{equation}
\mathrm{DCG@}K(q)=\sum_{k=1}^{K}\frac{2^{\mathrm{rel}_q(k)}-1}{\log_2(k+1)}.
\end{equation}
Since relevance is binary, the ideal discounted cumulative gain is
\begin{equation}
\mathrm{IDCG@}K(q)=
\sum_{k=1}^{\min(K,|R_q|)}
\frac{1}{\log_2(k+1)}.
\end{equation}
The normalized discounted cumulative gain is then
\begin{equation}
\mathrm{NDCG@}K(q)=\frac{\mathrm{DCG@}K(q)}{\mathrm{IDCG@}K(q)}.
\end{equation}

The MAP and NDCG values reported in the main paper are averaged over the five queries. The non-trivial match (NtM) is treated as a single binary-relevant segment for these metrics, whereas its additional pedagogical credit is used only in the top-10 pedagogical analysis.

\subsection{Generated Cannonball Illustrations}
\label{app:generated_cannonball}

\begin{figure}[t]
    \centering
    \begin{minipage}[b]{0.6\textwidth}
        \begin{subfigure}[b]{0.48\textwidth}
            \centering
            \includegraphics[height=2.5cm, keepaspectratio]{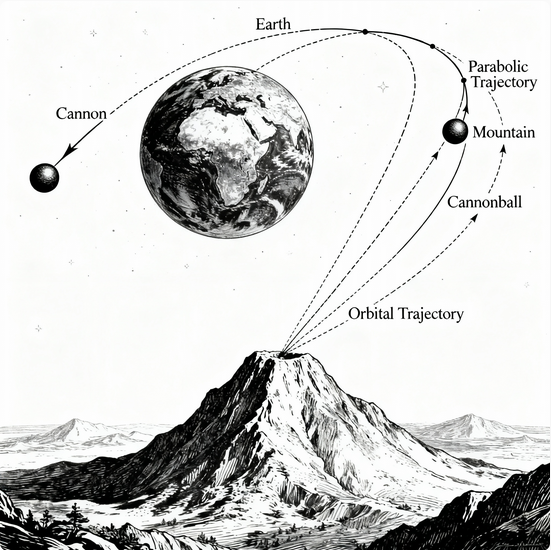}
            \caption{DALL-E (Jan., 2026)}
        \end{subfigure}
        \begin{subfigure}[b]{0.48\textwidth}
            \centering
            \includegraphics[height=2.5cm, keepaspectratio]{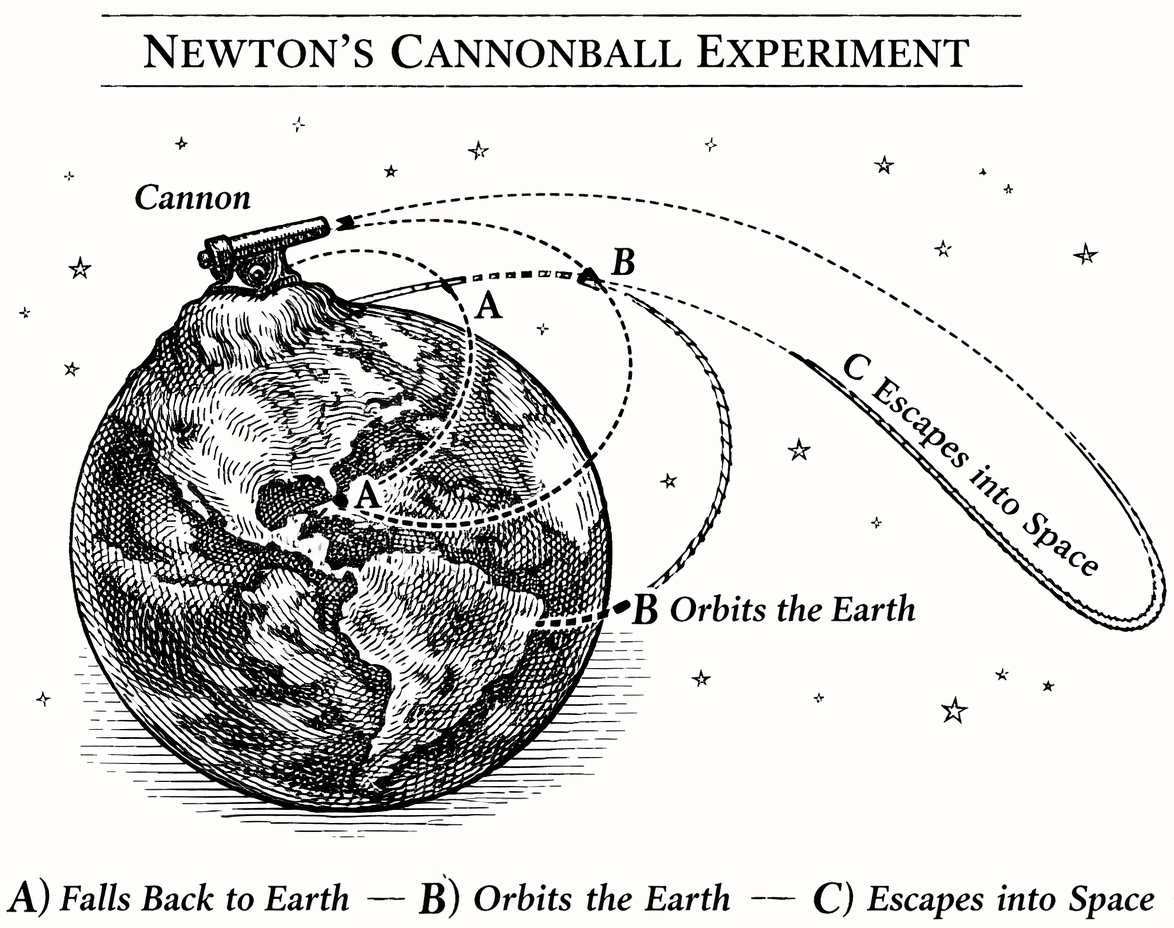}
            \caption{ChatGPT Image (Jan., 2026)}
        \end{subfigure}

        \vspace{2mm}

        \begin{subfigure}[b]{1\textwidth}
            \centering
            \includegraphics[height=3cm, keepaspectratio]{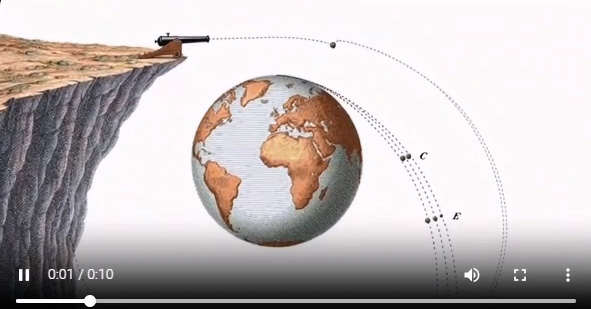}
            \caption{Sora2 (Jan., 2026)}
        \end{subfigure}
    \end{minipage}
    \begin{minipage}[b]{0.35\textwidth}
        \centering
        \begin{subfigure}[b]{0.8\textwidth}
            \centering
            \includegraphics[height=6.2cm, keepaspectratio]{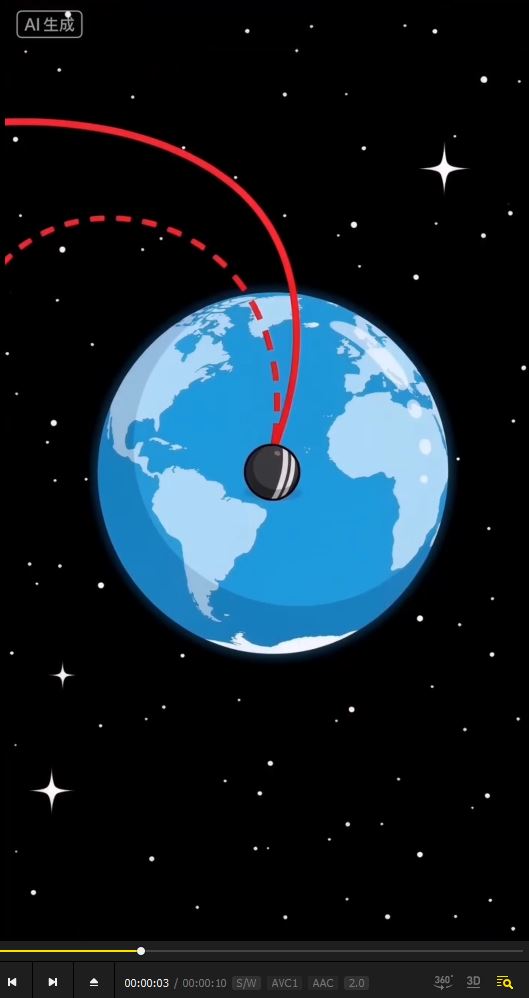}
            \caption{Seedance 2.0 (Feb., 2026)}
        \end{subfigure}
    \end{minipage}

    \caption{Generated depictions of Newton's cannonball thought experiment. Subfigures (a)--(c) share the prompt: ``Newton's cannonball thought experiment, Earth in space, parabolic trajectories, scientific diagram, vintage engraving style, clear labels, white background''. Subfigure (d) uses the Chinese prompt ``Newton's cannonball thought experiment''.}
    \label{fig:complex}
\end{figure}

Fig.~\ref{fig:complex} illustrates why retrieving scientifically produced documentary footage remains valuable even when generative models can synthesize diagrams or animations. The generated examples may be visually plausible, but they can contain physical inconsistencies, ambiguous trajectories, or prompt-intent deviations. In contrast, concept-to-example retrieval aims to locate vetted documentary evidence that can be used directly in teaching.

\subsection{Valid loss curve}
\begin{figure}[hbtp]
\centering
\includegraphics[width=\linewidth,height=0.3\textheight,keepaspectratio]{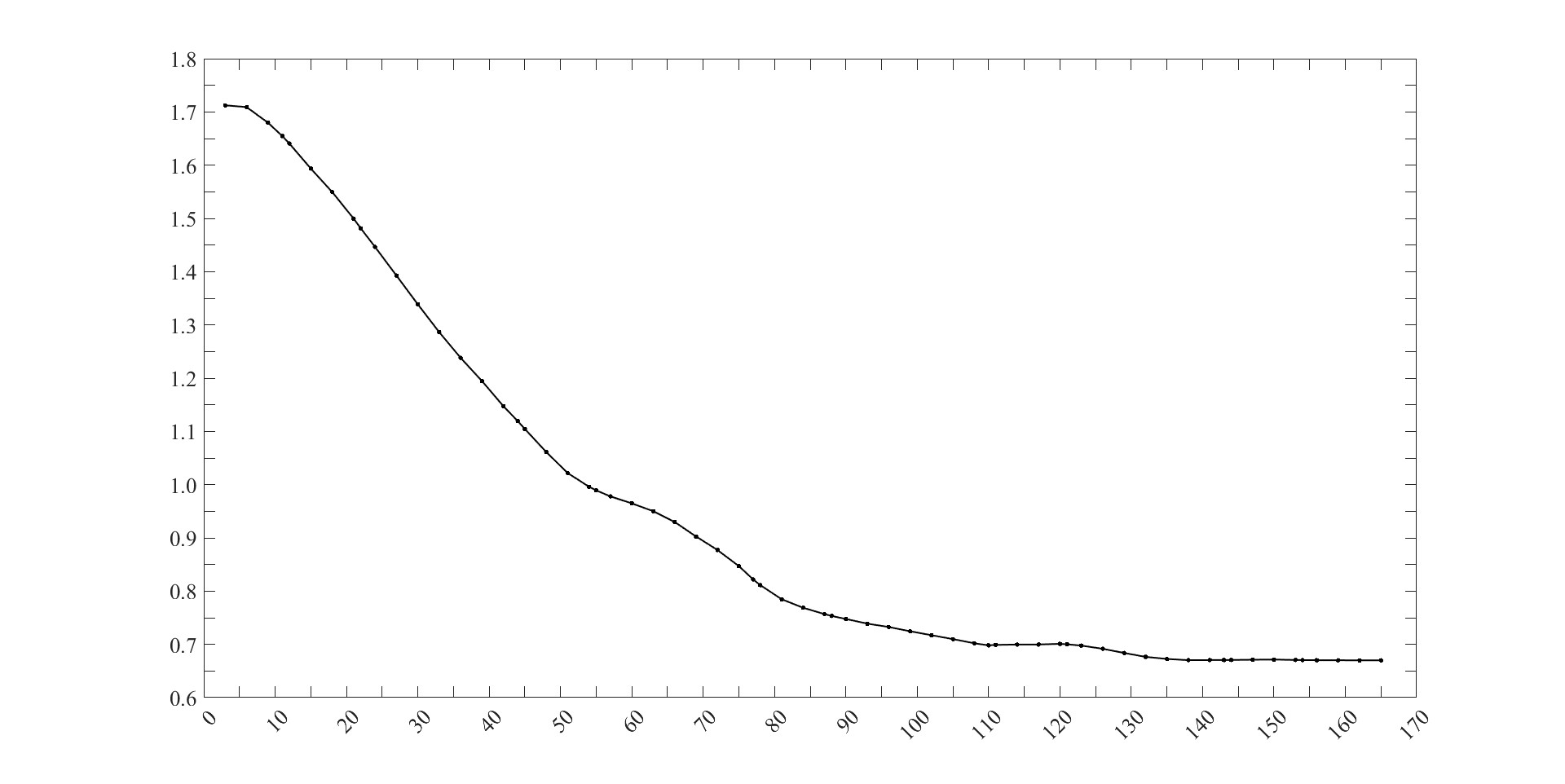}
\caption{\sethlcolor{red} Valid loss curve}
\label{fig:ValidLoss}
\end{figure}






\end{document}